\documentclass[10pt,twocolumn,letterpaper]{article}

\usepackage{wacv}
\usepackage{times}
\usepackage{epsfig}
\usepackage{graphicx}
\usepackage{amsmath}
\usepackage{amssymb}
\usepackage{booktabs}

\usepackage{multirow}
\usepackage{ulem}
\usepackage{capt-of}
\usepackage{xcolor}
\usepackage{colortbl}
\usepackage[accsupp]{axessibility}

\def\wacvPaperID{99} 

\wacvalgorithmstrack   

\wacvfinalcopy 

\usepackage[pagebackref=true,breaklinks=true,colorlinks,bookmarks=false]{hyperref}

\begin{document}

\title{Simultaneous Acquisition of High Quality RGB Image and Polarization Information using a Sparse Polarization Sensor}

\author{Teppei Kurita \qquad Yuhi Kondo \qquad Legong Sun \qquad Yusuke Moriuchi\\
Sony Group Corporation\\
{\tt\small \{Teppei.Kurita, Yuhi.Kondo, Legong.Sun, Yusuke.Moriuchi\}@sony.com}\\
\small \url{https://github.com/sony/polar-densification}
}

\twocolumn[{
	\renewcommand\twocolumn[1][]{#1}
	\maketitle
	\begin{center}
		\vspace{-0.7cm}
        \begin{tabular}[b]{c}
          \includegraphics[width=1.4in]{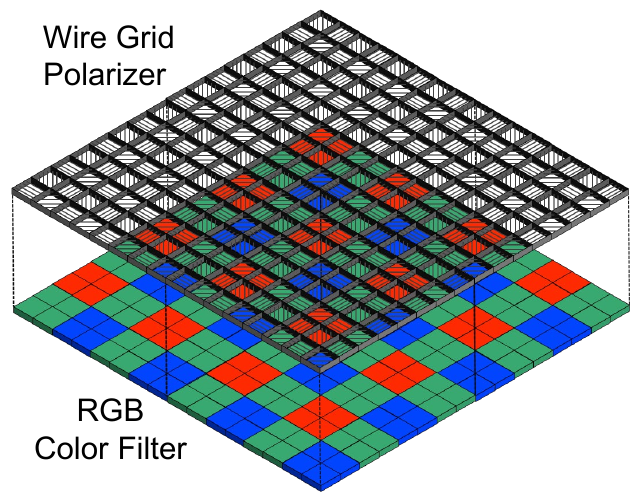}\\
          \scriptsize Conventional polarization sensor
        \end{tabular}
		\hspace{-0.45cm}
        \begin{tabular}[b]{c}
          \includegraphics[height=1.3in]{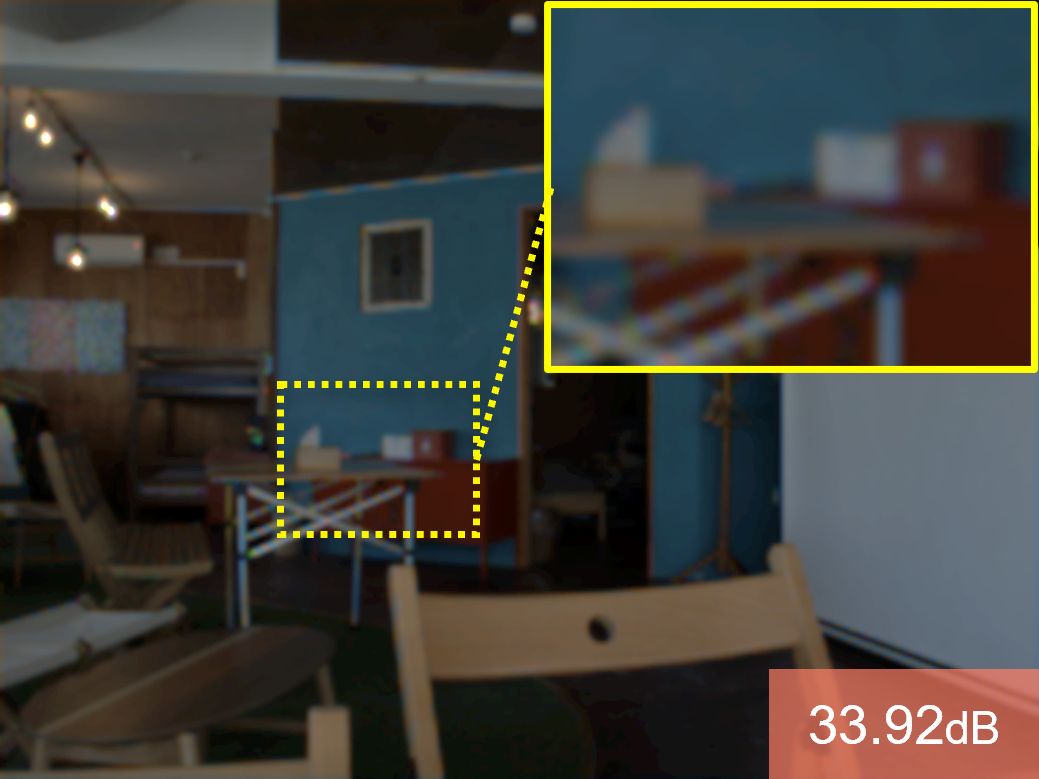}
        \end{tabular}
		\hspace{-0.55cm}
        \begin{tabular}[b]{c}
          \includegraphics[height=1.3in]{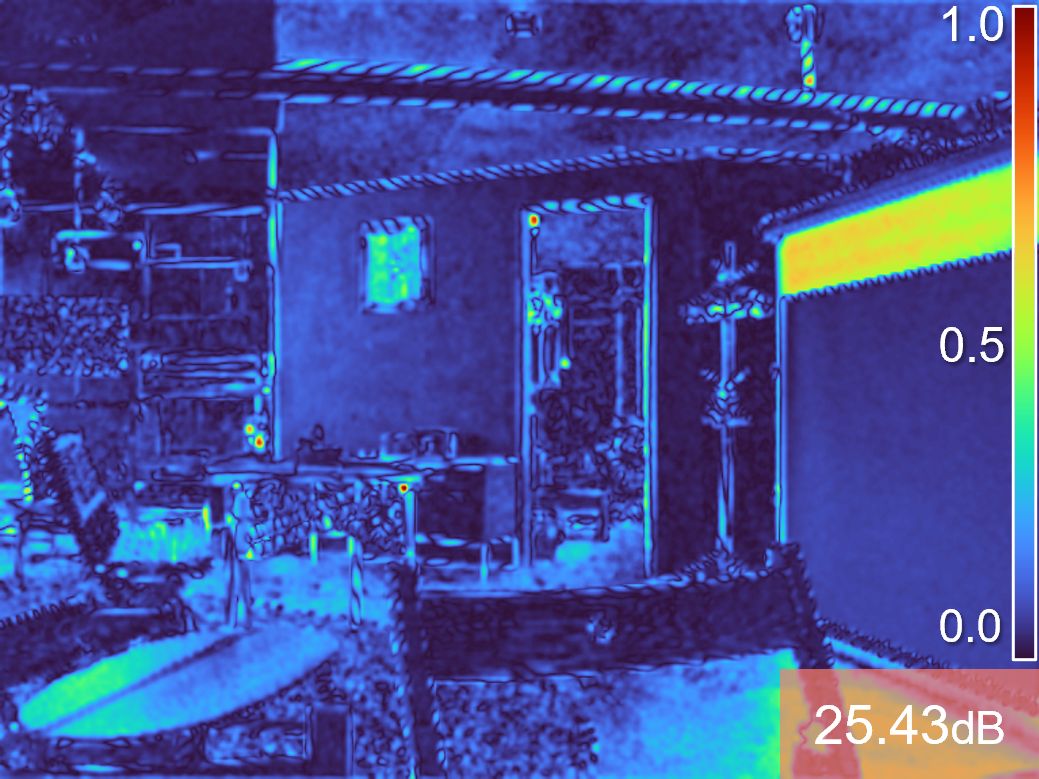}
        \end{tabular}
		\hspace{-0.55cm}
        \begin{tabular}[b]{c}
          \includegraphics[height=1.3in]{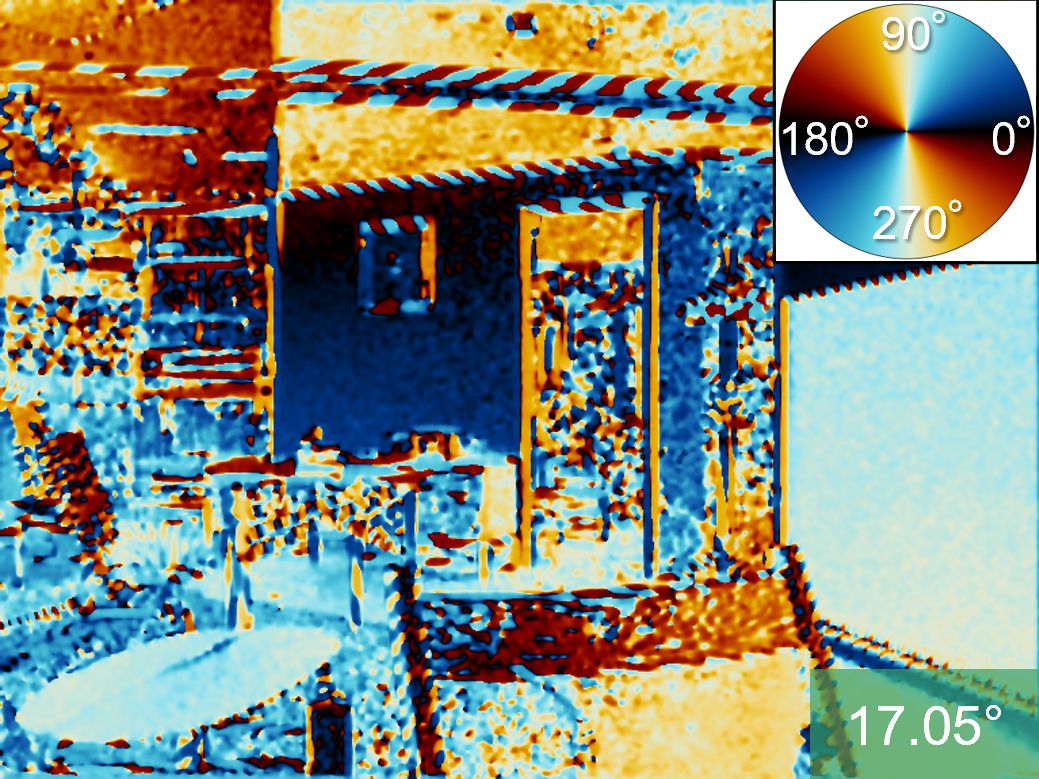}
        \end{tabular}\\
		\vspace{-0.1cm}
        \begin{tabular}[b]{c}
          \includegraphics[width=1.4in]{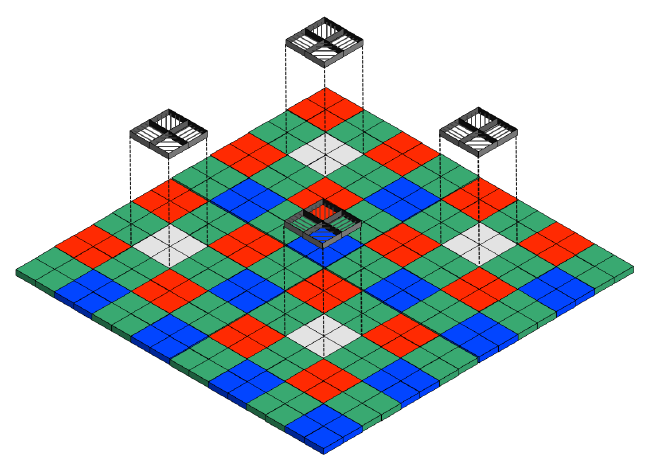}\\
          \scriptsize Sparse polarization sensor \vspace{-0.2cm}\\
          \scriptsize + our compensation\\
          \small (a) Sensor array
        \end{tabular}
		\hspace{-0.45cm}
        \begin{tabular}[b]{c}
          \includegraphics[height=1.3in]{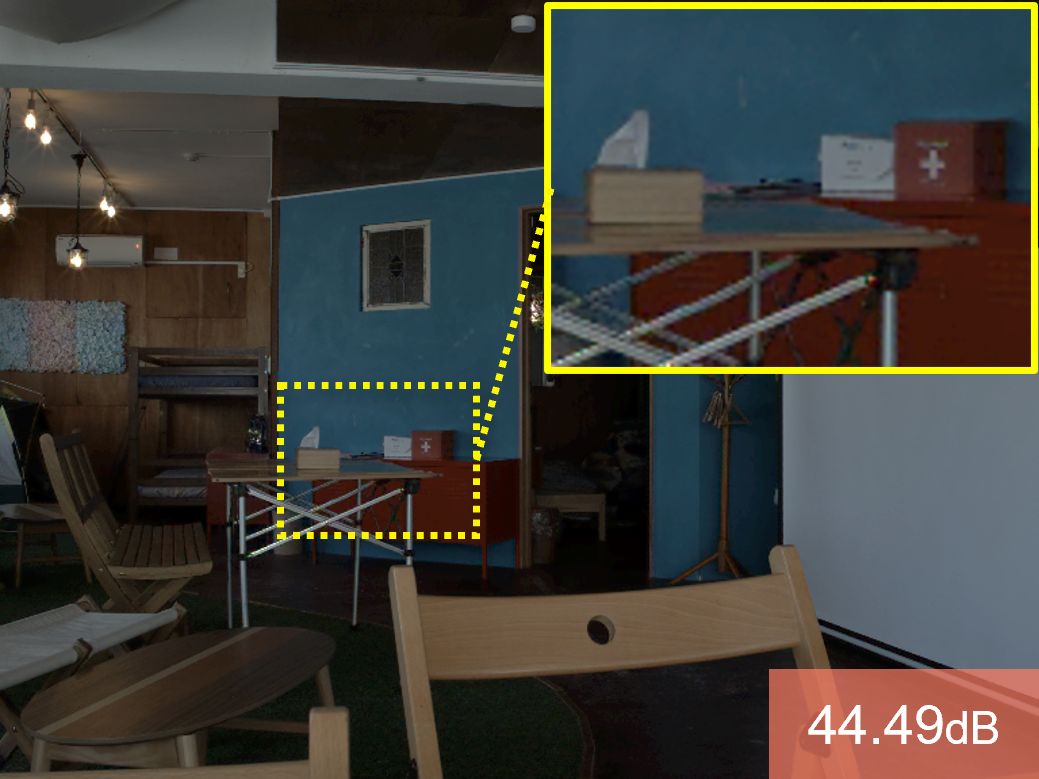}\\
          \small (b) RGB
        \end{tabular}
		\hspace{-0.55cm}
        \begin{tabular}[b]{c}
          \includegraphics[height=1.3in]{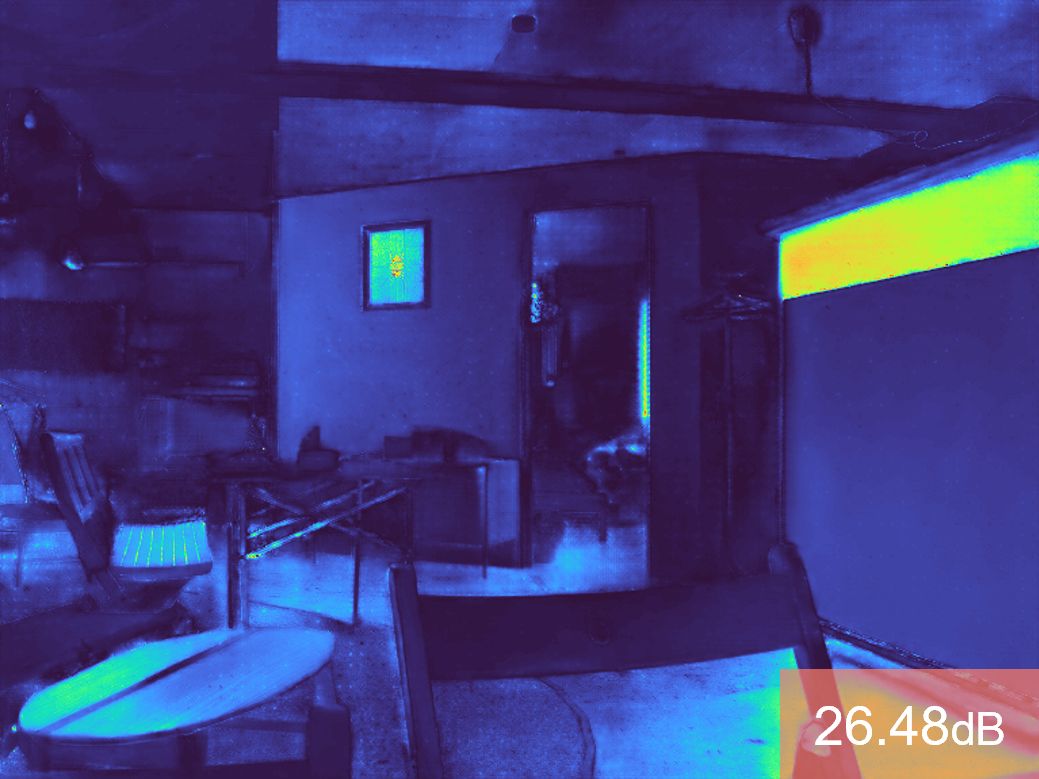}\\
          \small (c) DoLP
        \end{tabular}
		\hspace{-0.55cm}
        \begin{tabular}[b]{c}
          \includegraphics[height=1.3in]{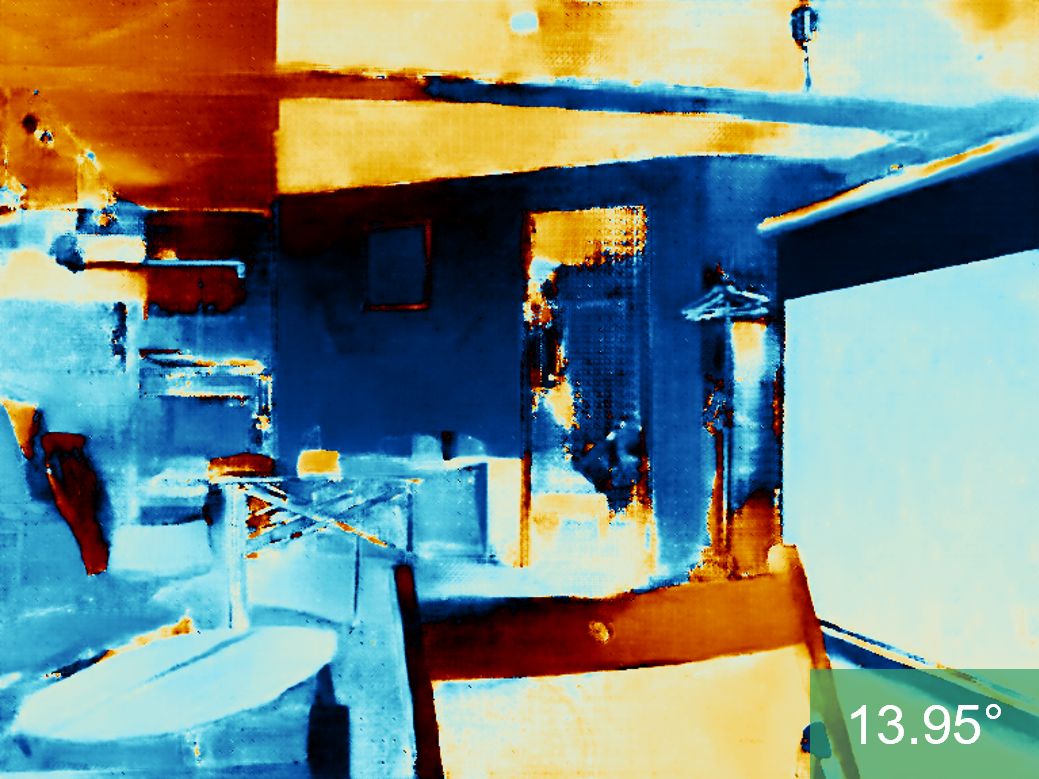}\\
          \small (d) AoLP
        \end{tabular}
		\captionof{figure}{
		\textbf{Compared to conventional polarization sensors, combining a sparse polarization sensor with a compensation method can enable more accurate RGB images and polarization information to be obtained.} The outputs of the conventional polarization sensor (top) and our proposed sensor (bottom). The RGB and degree of linear polarization (DoLP) values indicate the peak signal-to-noise ratio (PSNR) (the larger the better) and the angle of linear polarization (AoLP) values indicate angular error (the smaller the better).
		}
	\label{fig:teaser}
	\end{center}
}]

\thispagestyle{empty}

\begin{abstract}
\vspace{-0.4cm}
This paper proposes a novel polarization sensor structure and network architecture to obtain a high-quality RGB image and polarization information.
Conventional polarization sensors can simultaneously acquire RGB images and polarization information, but the polarizers on the sensor degrade the quality of the RGB images.
There is a trade-off between the quality of the RGB image and polarization information as fewer polarization pixels reduce the degradation of the RGB image but decrease the resolution of polarization information.
Therefore, we propose an approach that resolves the trade-off by sparsely arranging polarization pixels on the sensor and compensating for low-resolution polarization information with higher resolution using the RGB image as a guide.
Our proposed network architecture consists of an RGB image refinement network and a polarization information compensation network.
We confirmed the superiority of our proposed network in compensating the differential component of polarization intensity by comparing its performance with state-of-the-art methods for similar tasks: depth completion.
Furthermore, we confirmed that our approach could simultaneously acquire higher quality RGB images and polarization information than conventional polarization sensors, resolving the trade-off between the quality of RGB images and polarization information.
The baseline code and newly generated real and synthetic large-scale polarization image datasets are available for further research and development.
\end{abstract}

\vspace{-0.1cm}
\section{Introduction}
\begin{figure*}
\begin{center}
\includegraphics[width=\linewidth]{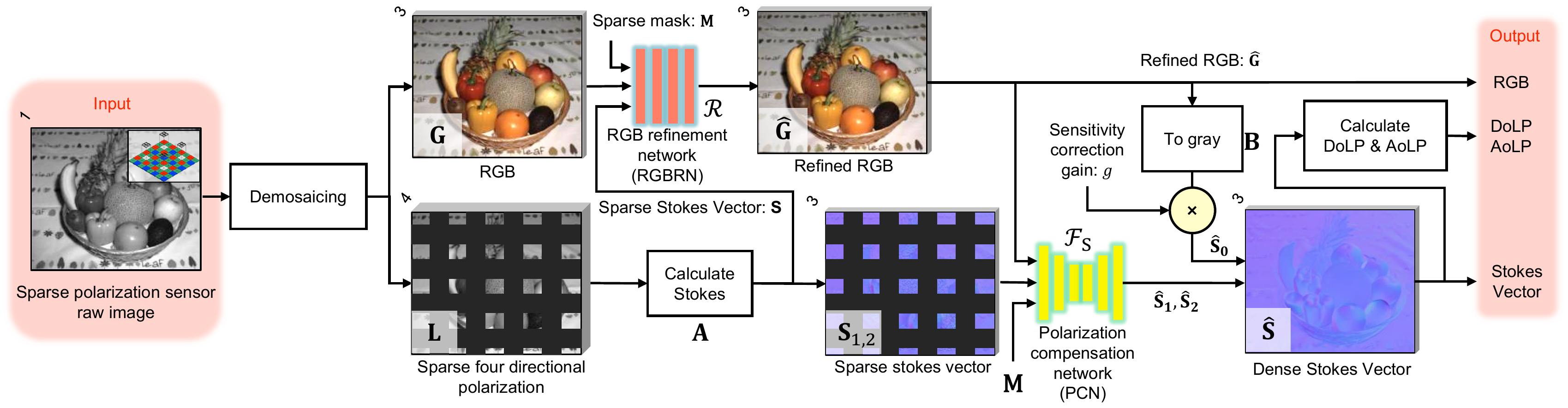}\\
\end{center}
\vspace{-0.5cm}
\caption{\textbf{Proposed stokes network architecture (SNA).}
The architecture consists of a refinement network for RGB images and compensation network for polarization information.
}
\label{fig:network}
\vspace{-0.4cm}
\end{figure*}
The polarization of light represents the orientation of the oscillations of light waves. It can be used to reveal light transport effects~\cite{baek2021polarimetric, tanaka2020polarized} such as shape~\cite{atkinson2005, atkinson2006}, transparency~\cite{kalra2020deep, mei2022glass, miyazaki2004transparent}, and scattering~\cite{treibitz2008active, zhang2019polarimetric}.
In recent years, polarization sensors \cite{yamazaki2016four, gruev2010}, which can simultaneously acquire the RGB image and polarization information in a single shot by placing the polarizer above the photodiode of the image sensor, as shown at the top in Fig.~\ref{fig:teaser} (a), have become widely used. This widespread implementation has enabled many applications such as shape estimation~\cite{ghosh2011multiview,ngo2015shape,kadambi2015polarized, qin2020u2, baek2018simultaneous, zhu2019depth, zhao2020polarimetric, fukao2021polarimetric, ichikawa2021shape, zou20203d}, reflection removal~\cite{lyu2019reflection, lei2020polarized, li2020reflection, deschaintre2021deep}, and so on~\cite{tzabari2020polarized, ting2021deep, li2020full, cui2019polarimetric, yang2018polarimetric, sturzl2017lightweight}.
The acquisition of polarization information is achieved by polarizers placed on the sensor. However, the sensor sensitivity is reduced due to the reduction in light intensity caused by the polarizers.
In addition, an unpolarized component is necessary to generate an RGB image, requiring pixel binning (averaging) of polarization components in multiple directions. This reduces the spatial resolution as shown at the top in Fig.~\ref{fig:teaser} (b) (on the other hand, sensitivity is improved by binning).
Reducing the degradation of the RGB image by reducing the number of polarization pixels on the sensor, as shown at the bottom in Fig.~\ref{fig:teaser} (a), is feasible in during design and manufacturing, but the resolution of the polarization information is reduced as a side effect. 
\begin{figure}
\begin{center}
\includegraphics[width=\linewidth]{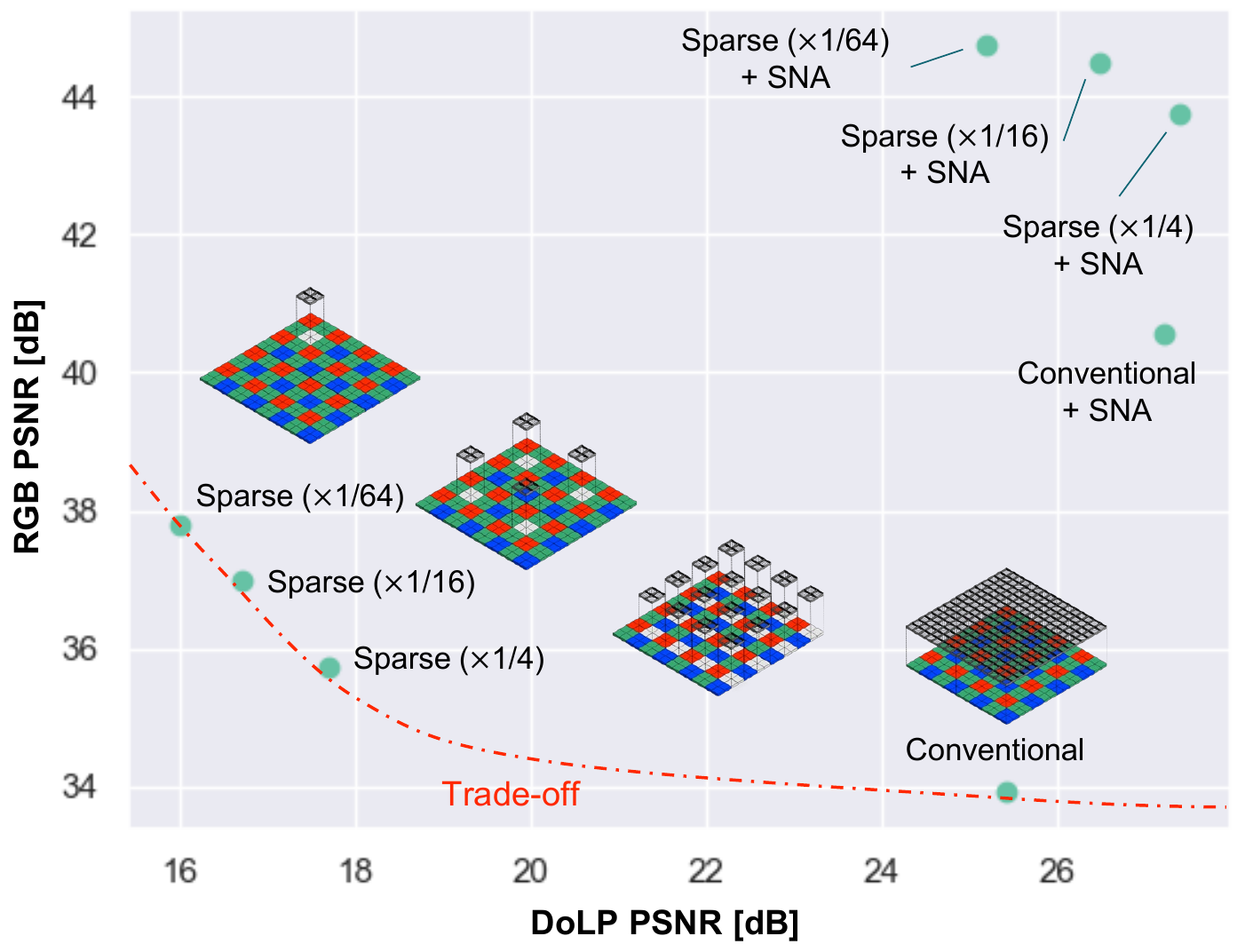}
\end{center}
\vspace{-0.5cm}
\caption{\textbf{Plot of PSNR for RGB and DoLP for each method.}
Our approach, SNA, resolves the trade-off between RGB image quality and polarization information quality.
}
\label{fig:trade-off}
\end{figure}

In this study, we propose an approach to resolve this trade-off between the quality of RGB images and polarization information by sparsely arranging polarization pixels on the sensor and compensating for low-resolution polarization information with higher resolution using the RGB image as a guide.
Our proposed stokes network architecture (SNA) consists of a refining network for RGB images and a compensation network for polarization information using the RGB images of the output of the refining network as a guide, as shown in Fig.~\ref{fig:network}.
The RGB image refinement network helps compensate for polarization information more effectively by correcting demosaicing artifacts and sparse pixels and cleaning up the RGB image used as a guide.
The compensation network for polarization information performs compensation only for polarization stokes components $\mathbf{S}_1$ and $\mathbf{S}_2$, which are differential components of polarization intensity. In contrast, the unpolarized component $\mathbf{S}_0$ is generated from the RGB image.
As only the difference component is learned, the polarization information can be compensated more stably than the method that compensates for polarization intensity in four directions.

In addition, large amounts of RGB and polarization data are not available to train networks.
Although the advent of polarization sensors has reduced the difficulty of acquiring polarization information, significant time and manpower are required to acquire large amounts of data.
We used polarization sensors to acquire real-world data and a polarization renderer ~\cite{kondo2020accurate} to generate a large synthetic dataset for training.
Our large synthetic dataset is generated using Houdini~\cite{houdini}, a 3D software program capable of procedural modeling for the automatic generation of objects. Hence the cost of acquiring a large amount of data is negligible.

We confirmed the superiority of the proposed network architecture by comparing its performance with state-of-the-art (SOTA) methods for similar tasks like depth completion and upsampling.
Furthermore, we confirmed that our approach could simultaneously acquire higher quality RGB images and polarization information than conventional polarization sensors, as shown at the bottom in Fig.~\ref{fig:teaser}. Specifically, we demonstrated a performance improvement of more than 10 dB and 1 dB in the PSNR for RGB and DoLP images, respectively.
Finally, we showed that resolving the trade-off between the quality of RGB images and polarization information is possible, as shown in Fig.~\ref{fig:trade-off}.
Due to the versatility of our compensation architecture, achieving higher quality output from conventional polarization sensors is also possible.
In summary, our contributions are as follows:
\begin{itemize}
\item An approach to resolve the trade-off between RGB image quality and polarization information quality by using a sparse polarization sensor and compensation method.
\item An effective network architecture for compensating captured polarization information with degradation.
\item Providing large-scale polarization datasets that are high-quality real and diverse synthetic.
\end{itemize}

\section{Related work}
Several previous studies have proposed denoising~\cite{tibbs2018denoising, tibbs2017noise, li2020learning} or demosaicing~\cite{mihoubi2018survey, qiu2019polarization, liu2020new} to obtain high-quality polarization information.
Few research focus on the trade-off between the quality of RGB images and polarization information from polarization sensors.
Our strategy of sparsely arranging polarization pixels to avoid the degradation of RGB image quality is derived from image sensors with sparsely arranged phase detection pixels~\cite{kobayashi2016low}.
In these sensors, a portion of the pixels is intentionally light-shielded for image phase detection auto-focus~\cite{chan2017enhancement}. However, because the RGB image quality is comparable to that of ordinary sensors, these sensors have become popular in recent years in cameras for cell phones and other applications.
Therefore, we hypothesized that even if polarization pixels are arranged sparsely, sufficient RGB image quality can be obtained as long as appropriate compensation is provided. We have accordingly conducted research on this hypothesis.

Sparse polarization information has low resolution and needs to be adequately compensated using the RGB image as a guide.
Similar problems are encountered during the super-resolution of hyperspectral images, and depth completion and upsampling.
Because hyperspectral images often have low spatial resolution due to the adverse effects of dense spectral sampling, a lot of research on fusing hyperspectral images with high-resolution RGB images to produce high-resolution hyperspectral images has been conducted.
However, many methods~\cite{lanaras2015hyperspectral, dian2018deep} use constraints related to the unique physical properties of hyperspectral images and are not versatile enough to be applied directly to other problems.
In addition, depth completion~\cite{hu2021penet, cheng2020cspn++, yang2019dense, zhao2021adaptive, liu2021fcfr,gu2021denselidar, qiu2019deeplidar} or upsampling~\cite{kim2021deformable, li2019joint} is the task of taking a sparse depth map or low-resolution depth image and transforming it into high-resolution by referencing to an RGB image. This method has been the subject of much research, particularly using neural networks.
As the demand for autonomous driving, augmented reality, gesture recognition, etc., has increased, these fields have become more competitive, and SOTA networks~\cite{he2021towards, park2020non, tang2020learning} have become more sophisticated. However, because these methods are specialized for depth estimation, they produce many artifacts when applied directly for polarization information compensation.
Hence, in this study, we propose a new network architecture suitable for compensating polarization information.

\section{Method}
This section details the structure of the proposed sensor, problem formulation, network architecture for obtaining high-quality RGB images and polarization information, and newly acquired and generated real-world and synthetic datasets.

\subsection{Sparse polarization sensor}
An example of a sparse polarization sensor is shown at the bottom in Fig.~\ref{fig:teaser} (a). This is a structure where four polarization pixels of different angles are arranged in an $8\times8$ area in a Quad Bayer array RGB sensor. 
The proportion of polarized pixels is $r$, where $r = 1/16$.
A white color filter is placed in the polarization pixel area to increase the sensitivity of the sensor to the visible light wavelength band.

\begin{figure}
\begin{center}
    \begin{tabular}[b]{c}
      \includegraphics[height=1.3in]{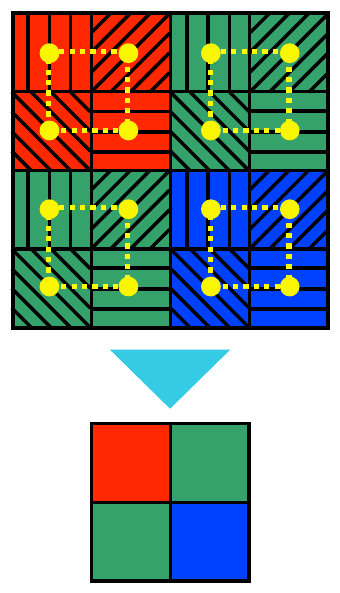}\\
      \small (a) Pixel binning
    \end{tabular}
	\hspace{-0.35cm}
    \begin{tabular}[b]{c}
      \includegraphics[height=1.28in]{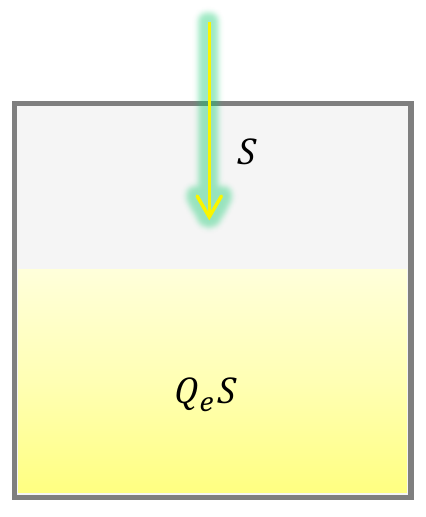}\\
      \small (b) Regular pixel
    \end{tabular}
	\hspace{-0.45cm}
    \begin{tabular}[b]{c}
      \includegraphics[height=1.28in]{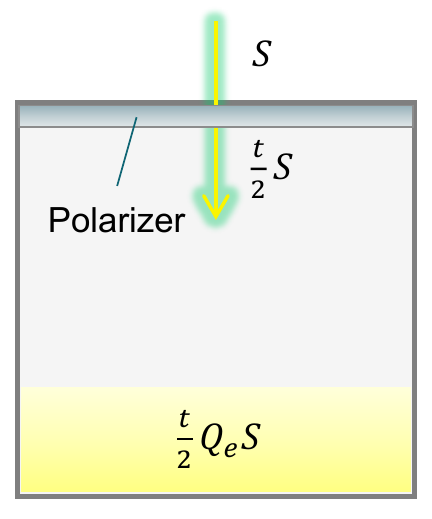}\\
      \small (c) Polarization pixel
    \end{tabular}
\end{center}
\vspace{-0.3cm}
\caption{\textbf{Pixel binning and sensitivity.}
(a) Pixel binning (averaging) in conventional polarization sensors.
Four pixels of the same color in a neighborhood are averaged to generate the unpolarized component, reducing resolution but increases sensitivity.
(b,c) Polarization pixel is less sensitive because the polarizer reduces the amount of light.
}
\label{fig:sensitivity_and_binning}
\end{figure}
\begin{figure}
\begin{center}
\small Resolution (conventional polarization sensor = 1.0)\\
\includegraphics[width=\linewidth]{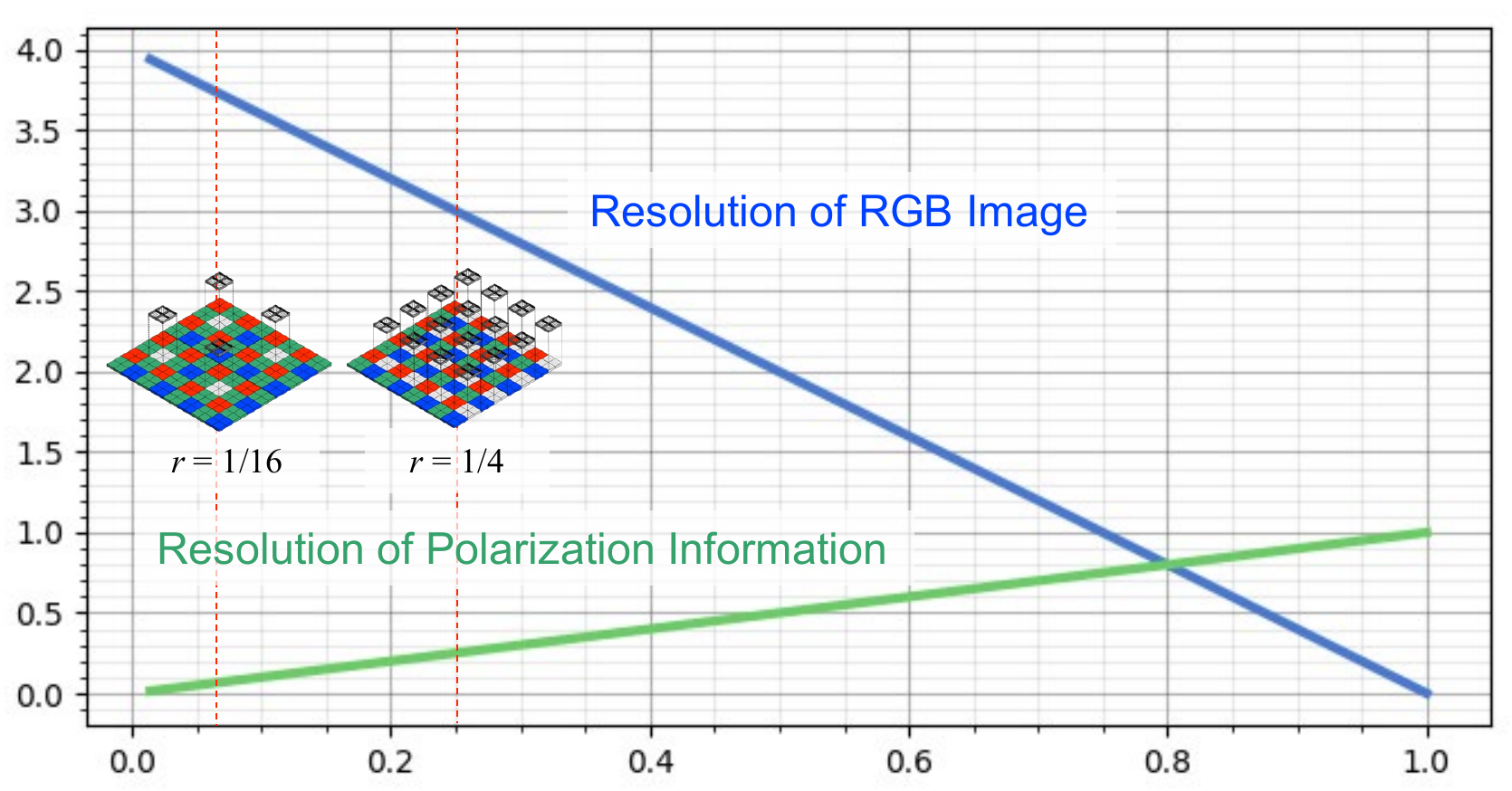}\\
\small Ratio of polarization pixels to total pixels: $r$
\end{center}
\vspace{-0.3cm}
\caption{\textbf{Resolution of sparse polarization sensors relative to conventional polarization sensors.}
There is a trade-off between the resolution of the RGB image and the resolution of polarization information.
}
\label{fig:resolution}
\vspace{-0.4cm}
\end{figure}

\vspace{-0.4cm}
\paragraph{Spatial resolution:}
To produce an RGB image using a conventional polarization sensor, pixel binning (averaging) of the four polarization angles must be done to create an unpolarized component, as shown in Fig.~\ref{fig:sensitivity_and_binning} (a).
Hence, if the total number of pixels is $N$, the number of RGB image pixels in a conventional polarization sensor would be $N/4$.
In contrast, the sparse polarization sensor can output regular pixels without binning, and the number of pixels would be $(1-r)N$.
Thus, the sparse polarization sensor has $4(1-r)$ times more pixels in the RGB image than the conventional polarization sensor, translating to 3.75 times more pixels when $r=1/16$.
As the sparse polarization sensor has fewer polarization pixels, the number of pixels for polarization information is $r$ times that of a conventional polarization sensor.
Hence, as shown in Fig.~\ref{fig:resolution}, the relationship between the resolution of the RGB image and the resolution of polarization information for the percentage of polarized pixels $r$ is a trade-off.

\vspace{-0.4cm}
\paragraph{Sensitivity (signal-to-noise ratio (SNR)):}
Conventional polarization sensors generate RGB images from polarized pixels, whereas sparse polarization sensors generate RGB images from regular pixels.
As shown in Fig.~\ref{fig:sensitivity_and_binning} (b,c), when unpolarized light enters a polarized pixel, the light intensity becomes $t/2$ times that of a regular pixel, where  
$t$ is the transmittance of the polarizer $(0\leq t \leq1)$.
For simplicity, assume that the noise factor, read noise and transmittance of the color filter are negligible, and the number of photons and quantum efficiency are $S$ and $Q_\mathrm{e}$, respectively.
Accordingly, the SNR of a regular pixel and polarized pixel is $\sqrt{Q_\mathrm{e}S}$ and $\sqrt{tQ_\mathrm{e}S/2}$, respectively.
When generating RGB images, the average SNR for the conventional polarization sensor is $\sqrt{2tQ_\mathrm{e}S}$ because it averages four polarization pixels, while the average SNR for the sparse polarization sensor is $\sqrt{Q_\mathrm{e}S}$.
Therefore, the sparse polarization sensor has $\sqrt{1/2t}$ times higher SNR for the RGB image than the conventional polarization sensor. When the transmittance is set to $t = 0.7$, the standard value for polarization sensors, the SNR is slightly lower at approximately 0.85 times.
Conversely, the sparse polarization sensors is slightly more sensitive to polarization information due to the white color filter.

\begin{table}
  \caption{\textbf{Characteristics of sparse polarization sensors compared to conventional polarization sensors.} $r$ is the ratio of the number of polarization pixels to the total number of pixels ($0 \leq r \leq 1$) and $t$ is the transmittance of the polarizer ($0 \leq t \leq 1$).}
  \centering
  \begin{tabular}{lcc}
    \toprule
      & RGB &  Polarization\\
    \midrule
    Resolution & \cellcolor{lime} $\times 4(1-r)$ & \cellcolor{pink} $\times r$ \\
    \midrule
    Sensitivity (SNR) & $\times \sqrt{1/2t}$ & Slightly better \\
    \bottomrule
  \end{tabular}
  \label{tbl:sparse_sensor}
\end{table}

Table~\ref{tbl:sparse_sensor} summarizes the characteristics of sparse polarization sensors compared to conventional polarization sensors.
The more sparse the polarization information, the better the resolution of the RGB image, whereas the resolution of the polarization information deteriorates.
Conversely, the sensitivity of the RGB image is slightly reduced and the sensitivity of the polarization information is slightly increased.
Regardless, these changes are trivial compared to the spatial resolution, on which the quality of the final RGB image and polarization information highly depend.

\subsection{Problem formulation}
\label{sec:problem_formulation}
The aim of this research is to generate a compensated stokes component $\hat{\mathbf{S}}=[\hat{\mathbf{S}}_0,\hat{\mathbf{S}}_1,\hat{\mathbf{S}}_2] \in \mathbb{R}^{M \times N \times 3}$ for a high-resolution RGB image $\mathbf{G} \in \mathbb{R}^{M \times N \times 3}$ and sparse four-polarization image $\mathbf{L}=[\mathbf{L}_{0}, \mathbf{L}_{45}, \mathbf{L}_{90}, \mathbf{L}_{135}] \in \mathbb{R}^{M \times N \times 4}$ filled with zeros except for the polarization pixels.
$M$ and $N$ are the height and width of the image, respectively.
As the polarization sensor cannot acquire the circular polarization component, $\mathbf{S}_3$ is omitted.
The transformation matrix $\mathbf{A}$ transforms each component of the four polarization angles into its corresponding stokes vector as follows:
\begin{equation}
\hspace{-0.1cm}
\mathbf{S}=
\begin{bmatrix}
\mathbf{S}_0\\
\mathbf{S}_1\\
\mathbf{S}_2\\
\end{bmatrix}
=\mathbf{A}\cdot\mathbf{L}=
\begin{bmatrix}
(\mathbf{L}_0+\mathbf{L}_{45}+\mathbf{L}_{90}+\mathbf{L}_{135})/4\\
(\mathbf{L}_0-\mathbf{L}_{90})/2\\
(\mathbf{L}_{45}-\mathbf{L}_{135})/2
\end{bmatrix}
\label{eq:stokes}
\end{equation}

A naive solution is to learn a mapping $\mathcal{F}_\mathrm{L}:\mathbf{L} \rightarrow \hat{\mathbf{L}}$ such that $\mathbf{G}$ is used as a guide to produce a high-quality four-polarization angle image $\hat{\mathbf{L}}$, and then convert $\hat{\mathbf{L}}$ to a stokes vector $\hat{\mathbf{S}}$.
This conversion is carried out as follows:
\begin{equation}
\hat{\mathbf{S}}=\mathbf{A}\cdot\mathcal{F}_L(\mathbf{L}, \mathbf{G}; \theta_{\mathcal{F}_L}),\\
\label{eqn:naive1}
\end{equation}
where $\theta_{\mathcal{F}_\mathrm{L}}$ is the set of learned weights.
However, this method does not preserve the minute relationship between the four polarization angles well and produces significant artifacts.

Another basic approach is to learn the mapping $\mathcal{F}_\mathrm{S}:\mathbf{S} \rightarrow \hat{\mathbf{S}}$ such that instead of compensating for the four polarization angles directly, $\mathbf{L}$ is transformed into a sparse stokes vector $\mathbf{S}$ to produce a high-quality stokes vector $\hat{\mathbf{S}}$, as follows:
\begin{equation}
\hat{\mathbf{S}}=\mathcal{F}_\mathrm{S}(\mathbf{A} \cdot \mathbf{L}, \mathbf{G}; \theta_{\mathcal{F}_\mathrm{S}}).\\
\label{eqn:naive2}
\end{equation}
In this approach, components such as $\mathbf{S}_1$ and $\mathbf{S}_2$, primarily low-frequency and minor values, are recovered with relatively good quality. Regardless, the $\mathbf{S}_0$ component, with many high-frequency textures, is significantly degraded.
Another problem is that the artifacts primarily caused by sparse polarization pixels in the RGB image $\mathbf{G}$ degrades compensation performance, particularly near the edges.

Therefore, we propose the generation of the $\mathbf{S}_0$ component from the RGB image and the mapping $\mathcal{F}_\mathrm{S}$ to be performed using only the difference (polarization) components of the stokes vector, $\mathbf{S}_1$ and $\mathbf{S}_2$.
The $\mathbf{S}_0$ component of the stokes vector represents the unpolarized component. Therefore, as long as the sensitivity difference between pixels is absorbed, a higher quality image can be generated from the higher resolution unpolarized RGB image without generating from polarized pixels.
We also correct the demosaicing artifacts in the RGB image to produce a higher quality RGB image $\hat{\mathbf{G}}$ to be used as a guide for polarization information compensation.
Such a process is formulated as follows:
\begin{equation}
\begin{aligned}
&\mathbf{S} = \mathbf{A} \cdot \mathbf{L}, \; \hat{\mathbf{G}}=\mathcal{R}(\mathbf{G}, \mathbf{S}, \mathbf{M}; \theta_{\mathcal{R}}),\\
&\hat{\mathbf{S}}_0=g \cdot \mathbf{B} \cdot \hat{\mathbf{G}},\\
&\hat{\mathbf{S}}_{1,2}=\mathcal{F}_\mathrm{S}(\mathbf{S}_1, \mathbf{S}_2, \hat{\mathbf{G}}, \mathbf{M}; \theta_{\mathcal{F}_\mathrm{S}}),\\
\end{aligned}
\label{eqn:proposed}
\end{equation}
where the mapping $\mathcal{R}:\mathbf{G} \rightarrow \hat{\mathbf{G}}$ compensates for RGB, $g$ is a gain to absorb the sensitivity difference between the unpolarized and polarized pixels, $\mathbf{B}$ is a transformation matrix that converts the pixels from RGB to grayscale, and $\mathbf{M}$ is a mask for polarized pixels (a binary image with 1 for pixels with and 0 for pixels without polarization).

\subsection{Network architecture}
\label{sec:network}
This study proposes an end-to-end network architecture, SNA, that compensates for the low-resolution polarization pixels using RGB images, as shown in Fig.~\ref{fig:network}.
First, taking the RAW data from the sparse polarization sensor as input, a demosaicing process is performed to separate data into a three-channel RGB image and a four-channel image with four sparse polarization angles.
The sparse four-polarization angle image is converted to a stokes vector by the transformation matrix $\mathbf{A}$.
Next, the RGB image refinement network corrects the demosaicing artifacts in the RGB image.
Then, using the refined RGB image and polarization components of the sparse stokes components $\mathbf{S}_{1,2}$ as inputs, the polarization components of the high-resolution stokes components are estimated by a compensation network of polarization information.
After the refined RGB image is grayscaled and gain applied to absorb sensitivity differences, high-resolution $\mathbf{S}_0$ component is calculated to obtain the final stokes vector and RGB image.
The proposed SNA provides a compensation of $\mathbf{S}_{1,2}$, which is one of the main contributions of this study.
Compared to the basic compensation architecture that uses four polarization intensity images that are not separated into $\mathbf{S}_{1,2}$, SNA can compensate efficiently under polarization constraints (Tab.~\ref{tbl:ablation}).

\vspace{-0.4cm}
\paragraph{RGB refinement network (RGBRN):}
Sparse polarization sensors have polarization pixels interspersed with regular RGB sensors.
Therefore, demosaicing may cause artifacts in that polarized pixel region.
Such demosaicing artifacts in RGB images can interfere with the compensation of polarization information.
Hence, we use a refinement network to clean up the RGB image.
The network architecture is based on \cite{ignatov2018wespe}, with a modification in residual learning to stabilize learning.
In addition to the RGB image, sparse polarization components $\mathbf{S}_{0,1,2}$ and a mask $\mathbf{M}$ are used as input to complement the information in the missing polarization pixels, thereby improving refinement.

\vspace{-0.4cm}
\paragraph{Polarization compensation network (PCN):}
\begin{figure}
\includegraphics[width=\linewidth]{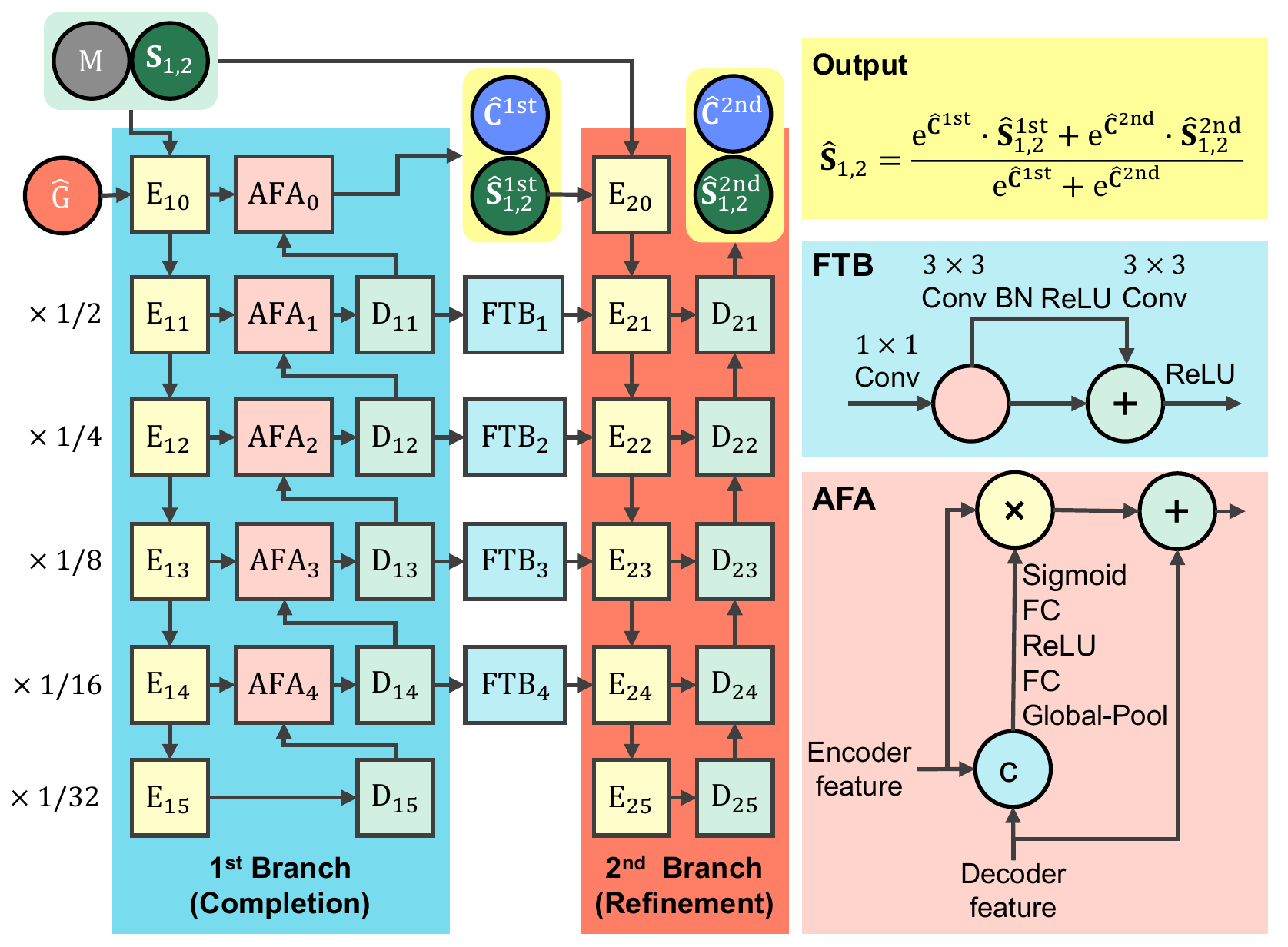}\\
\vspace{-0.4cm}
\caption{\textbf{Polarization compensation network (PCN).}
The network consists of one branch that complements sparse polarization information using the RGB image as a guide and a second branch that refines the obtained dense polarization information.
}
\label{fig:compensation}
\end{figure}
As previously mentioned, we compensate only for polarization information $\mathbf{S}_{1,2}$.
$\mathbf{S}_{1,2}$, the refined RGB image $\hat{\mathbf{G}}$, and mask $\mathbf{M}$ are taken as input, and compensation is performed by a two-branch network shown in Fig.~\ref{fig:compensation}, based on the ENet, the backbone in \cite{hu2021penet}.
The network first generates dense polarization information $\hat{\mathbf{S}}^{\mathrm{1st}}_{1,2}$, and confidence levels $\mathbf{C}^{\mathrm{1st}}$ as the first output, using the RGB image as a guide in the first branch. The branch is intended to complement the polarization information.
Next, the second blanch is used to refine the polarization information. The dense polarization information $\hat{\mathbf{S}}^{\mathrm{2nd}}_{1,2}$ and confidence level $\mathbf{C}^{\mathrm{2nd}}$ are generated as the second output using the first output $\hat{\mathbf{S}}^{\mathrm{1st}}_{1,2}$ and sparse $\mathbf{S}_{1,2}$ as inputs.
Finally, the first and second outputs are blended at their respective confidence levels $\hat{\mathbf{C}}$ to obtain polarization information $\hat{\mathbf{S}}_{1,2}$ as the final output.

Because ENet is intended to complement depth, it cannot be applied to complement polarization information. We expand the number of input and output channels to two and extract the ReLU just before the output to allow negative results.
The ENet has a direct skip connection of features from the first branch decoder to the same resolution layer of the second branch encoder.
However, features suitable for second branch refinement are not always generated by the first branch for completion purposes. This gap causes performance degradation.
Therefore, we use a feature transfer block (FTB) \cite{yu2018learning, li2018deep} to transfer the features generated by the first branch into a form suitable for the second branch.
In addition, the first branch of the ENet extracts RGB features through an encoder. It compensates $\mathbf{S}_{1,2}$ with the decoder, so different types of features are directly added with skip connections, causing a limitation in the expressive power of the network.
We apply an attention-based feature aggregation block (AFA) \cite{yu2018learning, li2018deep} that takes into account the global and visual features of the scene, assigns higher weights to important channels on the encoder side, and adds them to the decoder side features. This enables learning of more flexible representations in the first branch.

\vspace{-0.4cm}
\paragraph{Loss function:}
We use L1 loss for $\mathbf{S}_{1,2}$ and L2 loss for RGB $\mathbf{G}$ to train our network.
In the early stages of learning, the intermediate outputs, $\mathbf{S}_{1,2}^{\mathrm{1st}}$ and $\mathbf{S}_{1,2}^{\mathrm{2nd}}$, are also supervised, defined by
\begin{equation}
\begin{aligned}
&\mathcal{L}_\mathrm{S}(\hat{\mathbf{S}})=||\hat{\mathbf{S}}_{1,2}-\mathbf{S}_{1,2}^{\mathrm{gt}}||_1, \; \mathcal{L}_\mathrm{G}(\hat{\mathbf{G}})=||\hat{\mathbf{G}}-\mathbf{G}^{\mathrm{gt}}||_2\\
&\mathcal{L}=\mathcal{L}_\mathrm{S}(\hat{\mathbf{S}})+\lambda \{ \mathcal{L}_\mathrm{S}(\hat{\mathbf{S}}^{\mathrm{1st}})+\mathcal{L}_\mathrm{S}(\hat{\mathbf{S}}^{\mathrm{2nd}})\}+\mathcal{L}_\mathrm{G}(\hat{\mathbf{G}})
\end{aligned}
\label{eqn:loss}
\end{equation}
where $\mathbf{S}_{1,2}^{\mathrm{gt}}, \mathbf{G}^{\mathrm{gt}}$ are the ground truth, and $\lambda$ is the hyperparameter, a weight that decreases with the number of epochs.

\subsection{Dataset}
\label{sec:dataset}
\begin{figure}
\begin{center}
    \begin{tabular}[b]{c}
      \includegraphics[height=1.5in]{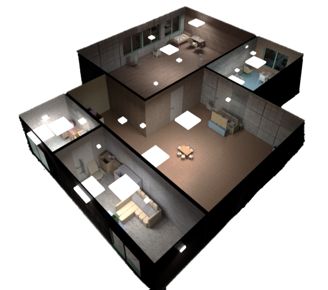}\\
      \small (a) Auto-generated floor and objects
    \end{tabular}
	\hspace{-0.35cm}
    \begin{tabular}[b]{c}
      \includegraphics[height=1.4in]{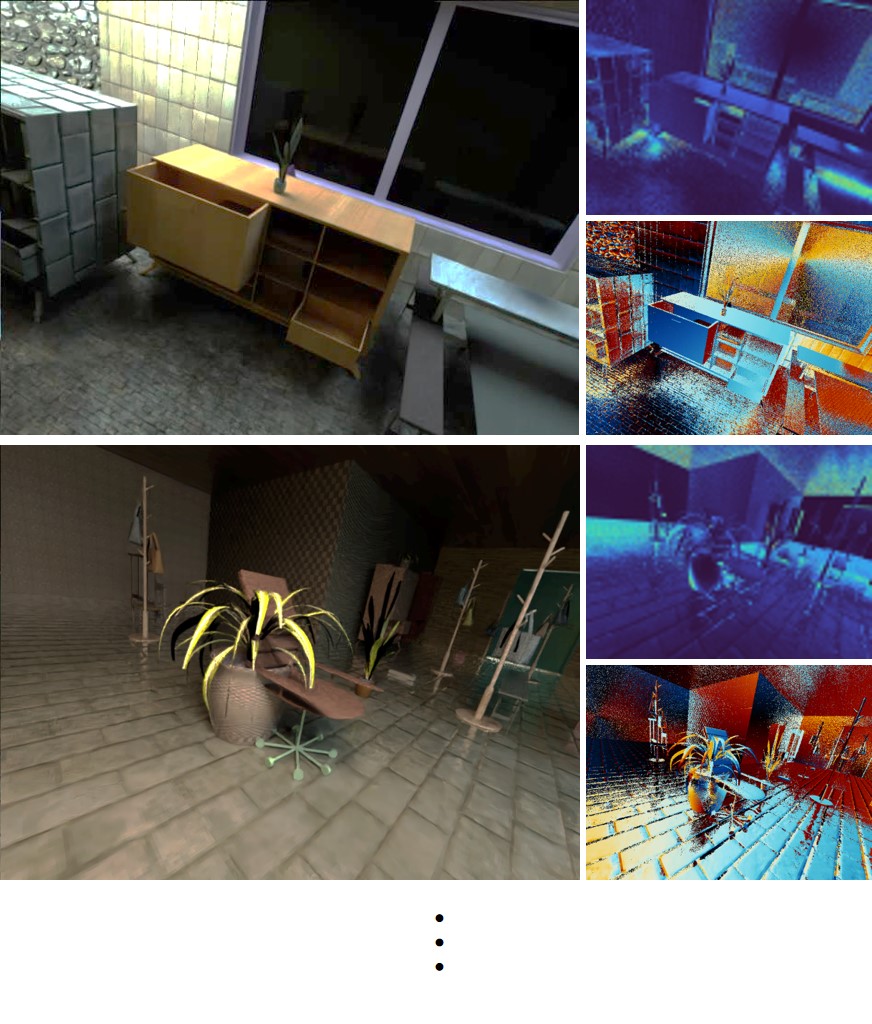}\\
      \small (b) Example of data
    \end{tabular}
\end{center}
\vspace{-0.3cm}
\caption{\textbf{
Automatically generated synthetic dataset.} (a) Floors and objects are generated procedurally. (b) Cameras are positioned randomly to acquire a wide variety of data at no cost.
}
\label{fig:synthetic_dataset}
\vspace{-0.3cm}
\end{figure}

\begin{table}
  \caption{Comparison among different polarization datasets}
  \centering
  \begin{tabular}{lccc}
    \toprule 
    Dataset  & Collection & Size & Resolution\\
    \midrule
    Ba \cite{ba2020deep} & Polarization camera & 263 & 1.3M\\
    Lei \cite{lei2022shape} & Polarization camera & 522 & 1.3M\\
    \multirow{2}{*}{Ono \cite{Ono_2022_CVPR}} & Polarization camera & 69 & 5.0M\\
    & Polarizer rotation & 13 & 4.3M\\
    \midrule
    \multirow{3}{*}{Ours}  & Synthetic & 11000 & 1.3M \& 0.4M\\
    & Polarization camera & 811 & 5.0M\\
    & Polarizer rotation & 238 & 20.0M\\
    \bottomrule
  \end{tabular}
  \label{tbl:dataset_comp}
  \vspace{-0.4cm}
\end{table}
We obtained real-world RGB and polarization information datasets using two methods.
One method uses a polarization sensor, which is less challenging to acquire data, but the data quality is inferior.
The other method is to rotate the polarizer in front of a regular RGB camera to acquire the data, which has higher quality but costs much more.
Hence, we generated a large synthetic dataset with our polarization renderer to compensate for these shortcomings.
Our approach uses Houdini to procedurally generate 3D objects and indoor scenes, as shown in Fig.~\ref{fig:synthetic_dataset}. Accordingly large amounts of data can be obtained at no cost.
We used rule-based floorplan generation \cite{camozzato2015procedural}, with object cameras randomly placed within the rules, and object textures and bump maps obtained from the Unreal Engine marketplace~\cite{unreal}.
Table~\ref{tbl:dataset_comp} shows a comparison of our real-world and synthetic datasets with existing datasets.
Our datasets outperformed both real-world and synthetic in their data size, and we acquired many high-resolution datasets.

\section{Experiments}
\subsection{Dataset and implementation details}
To evaluate our method, we performed numerous experiments on our dataset.
We used 10729 images (10000 synthetic and 729 real-world images acquired with an FLIR BFS-U3-51S5P-C color polarization sensor) for training and 1082 images (1000 synthetic and 82 real-world images) for validation.
For evaluation, we used 238 real-world images acquired by rotating the polarizer in front of the RGB camera FLIR BFS-U3-200S6C-C.
Raw data from a sparse polarization sensor with a resolution of $768 \times 576$ was generated by performing simulations on the dataset.
The transmittance of the polarizer was set to $t = 0.7$.
We added shot noise with a standard deviation of $F_{\mathrm{n}}\sqrt{S}$ to the real-world evaluation data based on a noise model taken from a real sensor, where $S$ is the pixel value and $F_{\mathrm{n}}$ is the noise factor.
We used $F_{\mathrm{n}}=0.72$ in our experiments unless otherwise stated, as it is the value for 0 dB analog gain on the real sensor.

Our network architecture was implemented in PyTorch on a PC equipped with an NVIDIA A100 GPU.
We performed 30 epochs of network training with a batch size of 5, an initial learning rate of 0.001 (gradual decay), and adopted the model of the epoch with the best validation results.
Using the loss defined in Eqn.~\ref{eqn:loss}, $\lambda = 0.2$ was set for the initial epoch and gradually reduced to 0.

\subsection{Assessment}
\begin{table}
  \caption{\textbf{Comparison with the basic methods (Sec.~\ref{sec:problem_formulation}).}
  $r$ is the percentage of polarized pixels.
  }
  \centering
  \begin{tabular}{l|lcccc}
    \toprule 
      $r$
      & Method
      & \begin{tabular}[c]{@{}l@{}}$\mathbf{S}_{0,1,2}$ \\RMSE $\downarrow$ \\ \scriptsize$[\times 10^{-3}]$ \end{tabular}
      & \begin{tabular}[c]{@{}l@{}}$\mathbf{S}_{1,2}$ \\RMSE $\downarrow$  \\ \scriptsize$[\times 10^{-3}]$ \end{tabular}
      & \begin{tabular}[c]{@{}l@{}}DoLP \\PSNR $\uparrow$ \\ \scriptsize [dB] \end{tabular}
      & \begin{tabular}[c]{@{}l@{}}AoLP \\Error $\downarrow$ \\ \scriptsize [$^\circ$]  \end{tabular}
      \\
    \midrule
    \multirow{3}{*}{$\frac{1}{4}$}
    & Eqn.~\ref{eqn:naive1} & 11.016 & 10.632 & 16.73 & 31.21\\
    & Eqn.~\ref{eqn:naive2} & 20.622 & 4.059 & 26.69 & 12.54\\
    & Ours & \textbf{4.881} & \textbf{3.824} & \textbf{27.41} & \textbf{12.36}\\
    \midrule
    \multirow{3}{*}{$\frac{1}{16}$}
    & Eqn.~\ref{eqn:naive1} & 9.103 & 8.955 & 19.00 & 27.49\\
    & Eqn.~\ref{eqn:naive2} & 27.562 & 4.353 & 25.40 & \textbf{13.57}\\
    & Ours & \textbf{4.707} & \textbf{4.151} & \textbf{26.48} & 13.95\\
    \midrule
    \multirow{3}{*}{$\frac{1}{64}$}
    & Eqn.~\ref{eqn:naive1} & 8.743 & 8.759 & 17.56 & 32.88\\
    & Eqn.~\ref{eqn:naive2} & 26.833 & 4.996 & 23.88 & 15.74\\
    & Ours & \textbf{5.032} & \textbf{4.801} & \textbf{25.19} & \textbf{15.56}\\
    \bottomrule
  \end{tabular}
  \label{tbl:eval_naive}
\end{table}
\begin{table}
  \caption{\textbf{Ablation study (Sec.~\ref{sec:network}).}
  Evaluation at $r=1/4$. 
  }
  \centering
  \begin{tabular}{lcccc}
    \toprule
    Operation
      & \begin{tabular}[c]{@{}l@{}}$\mathbf{S}_{0,1,2}$ \\RMSE $\downarrow$ \\ \scriptsize$[\times 10^{-3}]$ \end{tabular}
      & \begin{tabular}[c]{@{}l@{}}$\mathbf{S}_{1,2}$ \\RMSE $\downarrow$  \\ \scriptsize$[\times 10^{-3}]$ \end{tabular}
      & \begin{tabular}[c]{@{}l@{}}DoLP \\PSNR $\uparrow$ \\ \scriptsize [dB] \end{tabular}
      & \begin{tabular}[c]{@{}l@{}}AoLP \\Error $\downarrow$ \\ \scriptsize [$^\circ$]  \end{tabular}
      \\
    \midrule
    Baseline & 10.948 & 10.574 & 16.57 & 31.70\\
    + SNA & 10.328 & 4.046 & 26.94 & \textbf{12.31}\\
    + RGBRN & 4.974 & 3.979 & 26.75 & 13.43\\
    + FTB & 4.952 & 3.952 & 26.81 & 13.14\\
    + AFA (Ours) & \textbf{4.881} & \textbf{3.824} & \textbf{27.41} & 12.36\\
    \bottomrule
  \end{tabular}
  \label{tbl:ablation}
  \vspace{-0.3cm}
\end{table}
\begin{table}
  \caption{\textbf{
  Comparison of the results obtained after replacing PCN with depth completion and upsampling networks in our network architecture (Sec.~\ref{sec:network}).
  }
  Evaluation at $r=1/16$. 
  }
  \centering
  \begin{tabular}{lcccc}
    \toprule
      Method
      & \begin{tabular}[c]{@{}l@{}}$\mathbf{S}_{0,1,2}$ \\RMSE $\downarrow$ \\ \scriptsize$[\times 10^{-3}]$ \end{tabular}
      & \begin{tabular}[c]{@{}l@{}}$\mathbf{S}_{1,2}$ \\RMSE $\downarrow$  \\ \scriptsize$[\times 10^{-3}]$ \end{tabular}
      & \begin{tabular}[c]{@{}l@{}}DoLP \\PSNR $\uparrow$ \\ \scriptsize [dB] \end{tabular}
      & \begin{tabular}[c]{@{}l@{}}AoLP \\Error $\downarrow$ \\ \scriptsize [$^\circ$]  \end{tabular}
      \\
    \midrule
    UNet~\cite{ronneberger2015u} & 4.974 & 4.568 & 25.00 & 16.31\\
    U2Net~\cite{qin2020u2} & 5.537 & 5.224 & 23.90 & 19.11\\
    FDSR~\cite{he2021towards} & 5.128 & 4.837 & 25.12 & 15.48\\
    NLSPN~\cite{park2020non} & 4.905 & 4.470 & 23.97 & 20.20\\
    GuideNet~\cite{tang2020learning} & 4.859 & 4.390 & 25.59 & 15.62\\
    PCN (Ours) & \textbf{4.707} & \textbf{4.151} & \textbf{26.48} & \textbf{13.95}\\ \bottomrule
  \end{tabular}
  \label{tbl:network_comparison}
\end{table}
\begin{table}
  \caption{\textbf{Comparison of real-world (R) and synthetic (S) datasets (Sec.~\ref{sec:dataset}).}
  Evaluation at $r=1/4$. 
  }
  \centering
  \begin{tabular}{lccccc}
    \toprule
    Data
      & \begin{tabular}[c]{@{}l@{}}Train\\size\end{tabular}
      & \begin{tabular}[c]{@{}l@{}}$\mathbf{S}_{0,1,2}$ \\RMSE $\downarrow$ \\ \scriptsize$[\times 10^{-3}]$ \end{tabular}
      & \begin{tabular}[c]{@{}l@{}}$\mathbf{S}_{1,2}$ \\RMSE $\downarrow$  \\ \scriptsize$[\times 10^{-3}]$ \end{tabular}
      & \begin{tabular}[c]{@{}l@{}}DoLP \\PSNR $\uparrow$ \\ \scriptsize [dB] \end{tabular}
      & \begin{tabular}[c]{@{}l@{}}AoLP \\Error $\downarrow$ \\ \scriptsize [$^\circ$]  \end{tabular}
      \\
    \midrule
    R & 729 & 11.162 & 5.240 & 25.40 & 14.79\\
    S & 729 & 6.346 & 5.342 & 24.05 & 15.55\\
    S & 10000 & 5.137 & 4.093 & 26.58 & 13.11\\
    R+S & 10729 & \textbf{4.881} & \textbf{3.824} & \textbf{27.41} & \textbf{12.36}\\
    \bottomrule
  \end{tabular}
  \label{tbl:dataset_eval}
  \vspace{-0.4cm}
\end{table}
\begin{figure*}
\begin{center}
	\vspace{-0.7cm}
    \begin{tabular}[b]{c}
      \includegraphics[height=1in]{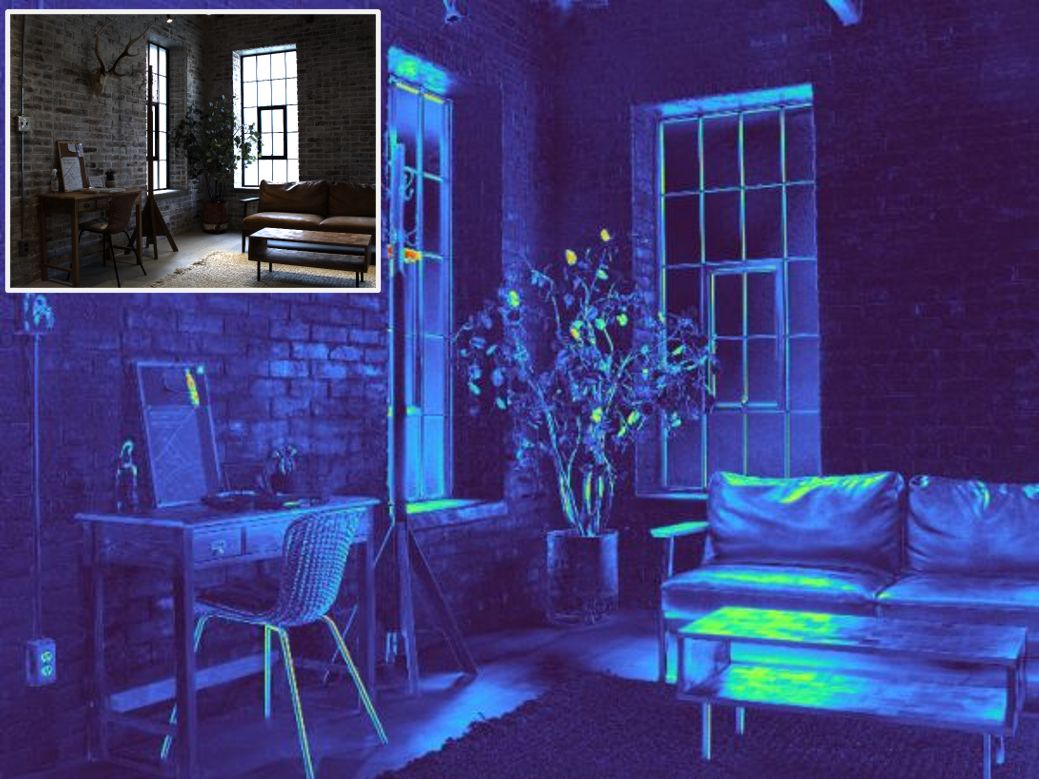}
    \end{tabular}
	\hspace{-0.58cm}
    \begin{tabular}[b]{c}
      \includegraphics[height=1in]{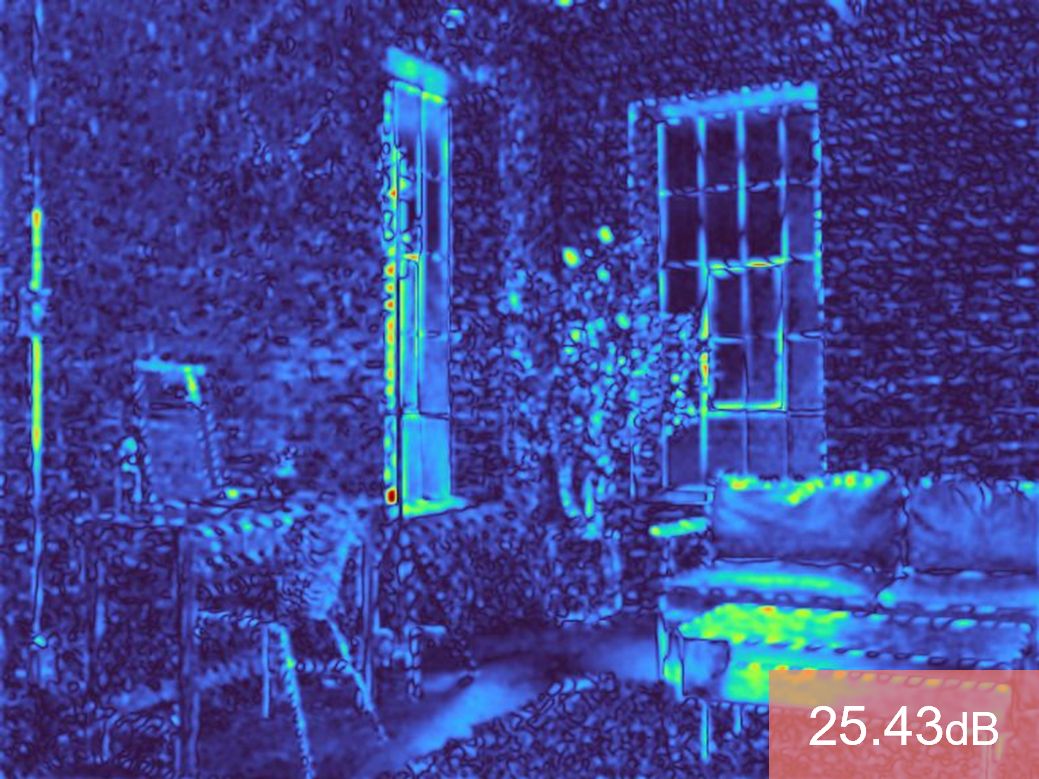}
    \end{tabular}
	\hspace{-0.58cm}
    \begin{tabular}[b]{c}
      \includegraphics[height=1in]{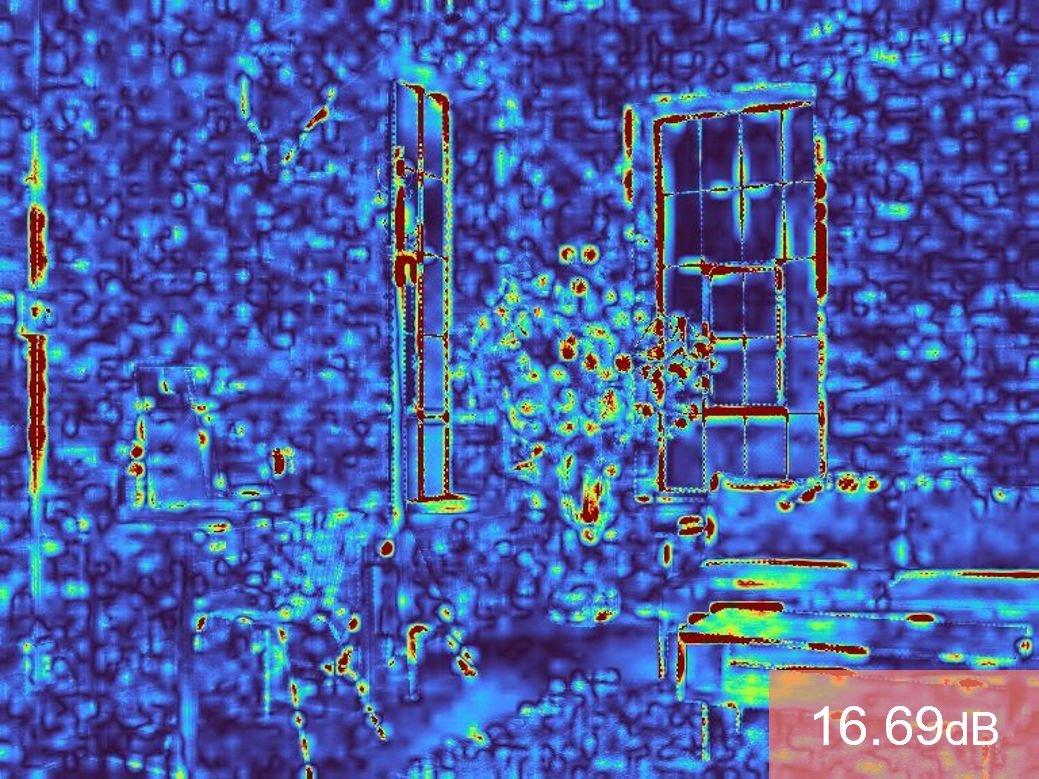}
    \end{tabular}
	\hspace{-0.58cm}
    \begin{tabular}[b]{c}
      \includegraphics[height=1in]{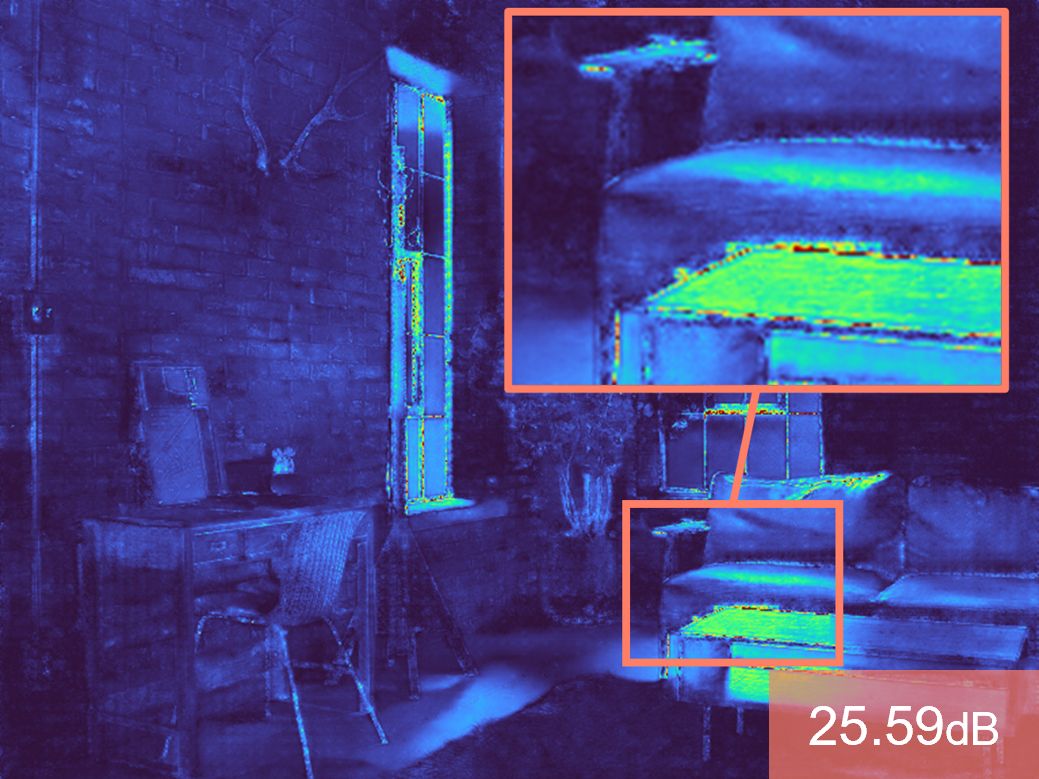}
    \end{tabular}
	\hspace{-0.58cm}
    \begin{tabular}[b]{c}
      \includegraphics[height=1in]{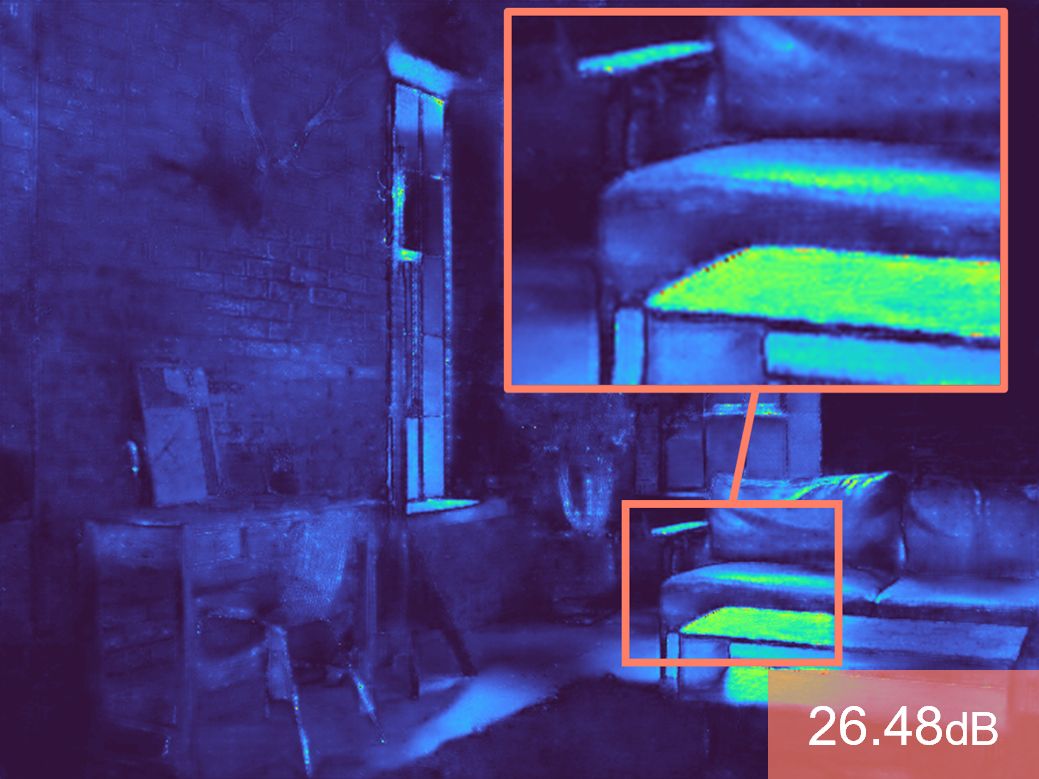}
    \end{tabular}\\
	\vspace{-0.14cm}
    \begin{tabular}[b]{c}
      \includegraphics[height=1in]{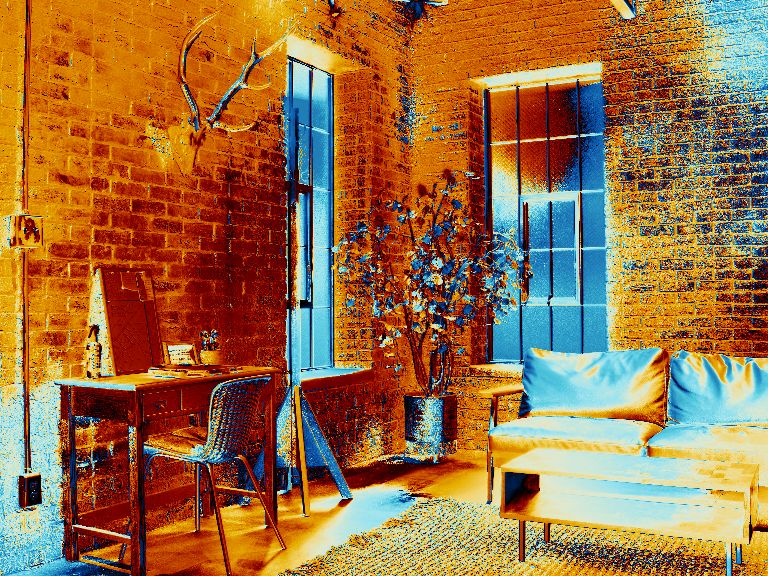}\\
      \small (a) Ground truth\\
      \small 
    \end{tabular}
	\hspace{-0.58cm}
    \begin{tabular}[b]{c}
      \includegraphics[height=1in]{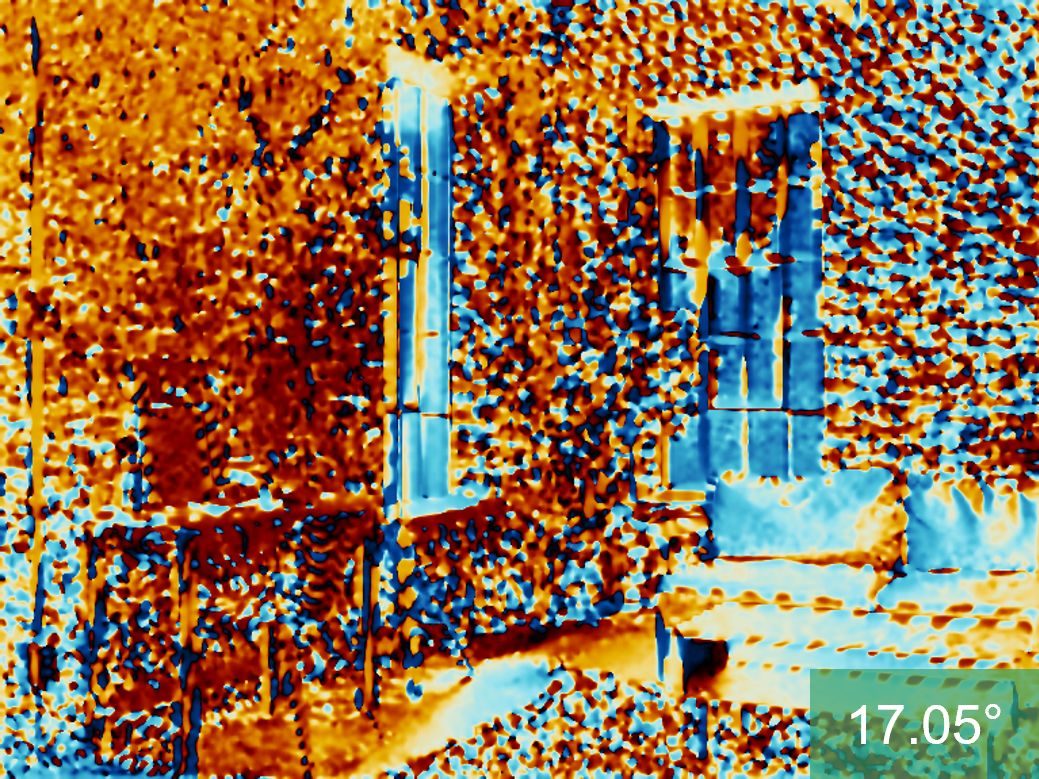}\\
      \small (b) Conventional\\
      \small
    \end{tabular}
	\hspace{-0.58cm}
    \begin{tabular}[b]{c}
      \includegraphics[height=1in]{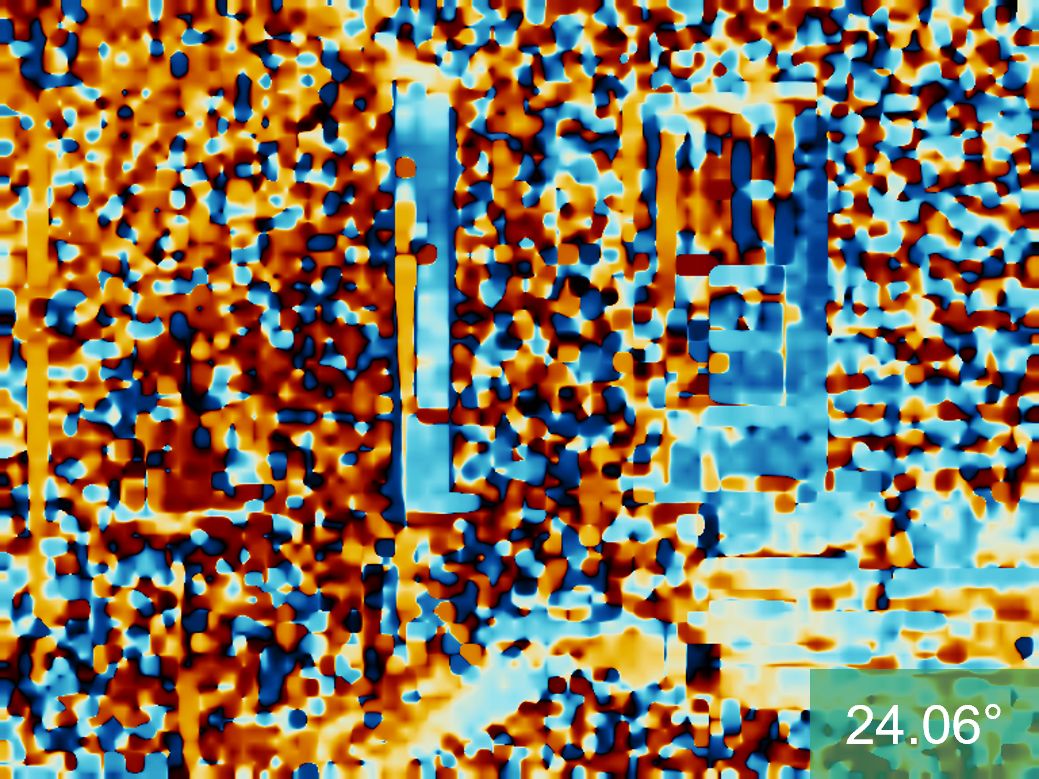}\\
      \small (c) Sparse + bilinear\\
      \small 
    \end{tabular}
	\hspace{-0.58cm}
    \begin{tabular}[b]{c}
      \includegraphics[height=1in]{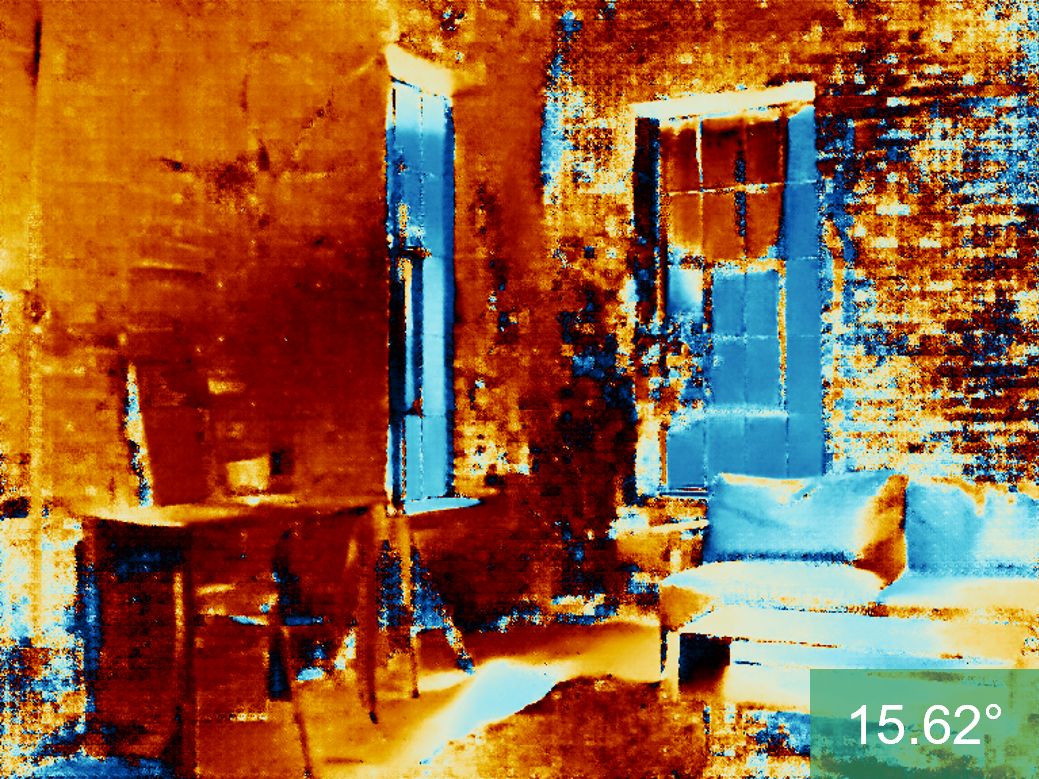}\\
      \small (d) Sparse + SNA\\
      \small with GuideNet \cite{tang2020learning}
    \end{tabular}
	\hspace{-0.58cm}
    \begin{tabular}[b]{c}
      \includegraphics[height=1in]{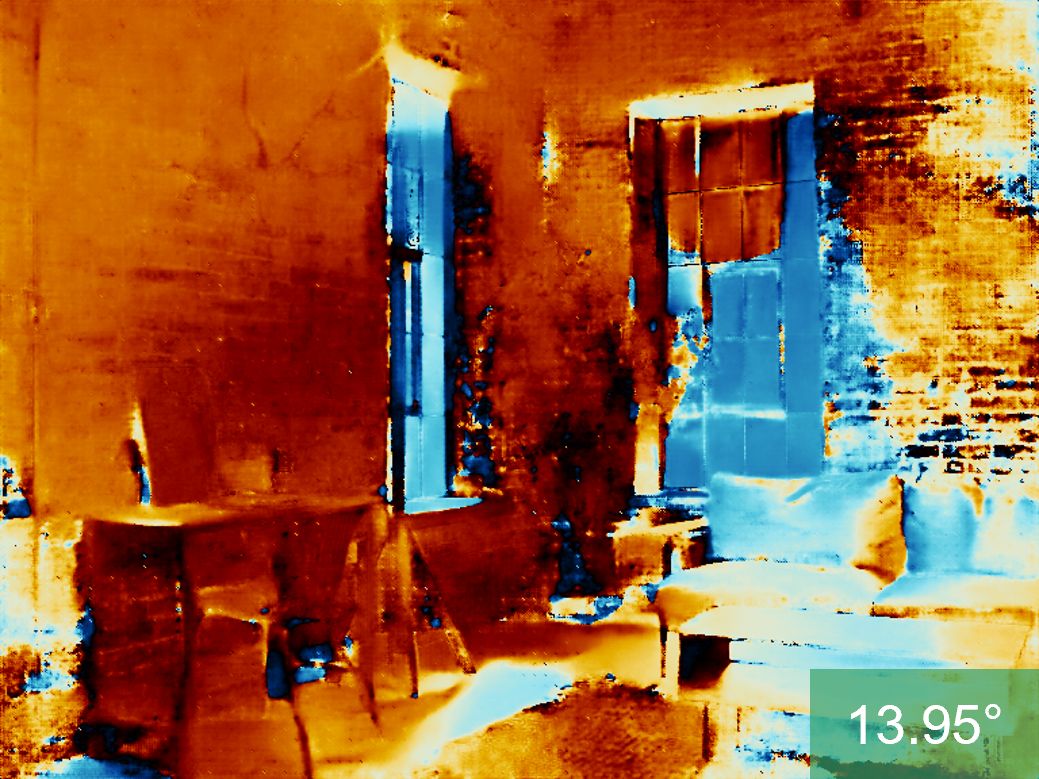}\\
      \small (e) Sparse + SNA\\
      \small with PCN (Ours)
    \end{tabular}
	\captionof{figure}{
	\textbf{Comparison with other methods.} Evaluation at $r=1/16$.
	The top is DoLP and the bottom is AoLP.
	(d) shows the results obtained after replacing our PCN with GuideNet in the proposed stokes network architecture (SNA) described in Sec.~\ref{sec:network}.
	}
\label{fig:result}
\end{center}
\vspace{-0.7cm}
\end{figure*}

\begin{table*}
  \caption{\textbf{Comprehensive evaluation of conventional and sparse polarization sensors, taking noise into account.}
  $F_{\mathrm{n}}$ is the noise factor (higher is noisier).
  }
  \centering
  \begin{tabular}{c|c|c|lcccccc}
    \toprule 
      $F_{\mathrm{n}}$
      & \begin{tabular}[c]{@{}c@{}}Polarization\\sensor\end{tabular}
      & $r$
      & Method
      & \begin{tabular}[c]{@{}l@{}}$\mathbf{S}_{0,1,2}$ \\RMSE $\downarrow$ \\ \scriptsize$[\times 10^{-3}]$ \end{tabular}
      & \begin{tabular}[c]{@{}l@{}}$\mathbf{S}_{1,2}$ \\RMSE $\downarrow$  \\ \scriptsize$[\times 10^{-3}]$ \end{tabular}
      & \begin{tabular}[c]{@{}l@{}}DoLP \\PSNR $\uparrow$ \\ \scriptsize [dB] \end{tabular}
      & \begin{tabular}[c]{@{}l@{}}AoLP \\Error $\downarrow$ \\ \scriptsize [$^\circ$]  \end{tabular}
      & \begin{tabular}[c]{@{}l@{}}RGB \\PSNR $\uparrow$ \\ \scriptsize [dB] \end{tabular}
      & \begin{tabular}[b]{@{}l@{}}RGB \\SSIM $\uparrow$ \end{tabular}
      \\
    \midrule
    \multirow{8}{*}{0.72}
    &\multirow{2}{*}{\begin{tabular}[c]{@{}l@{}}Conventional\end{tabular}}
    & \multirow{2}{*}{0}
    & - & 15.124 & 6.629 & 25.43 & 17.05 & 33.92 & 0.9251\\
    &&& Ours & 6.915 & 3.854 & 27.22 & \textbf{12.35} & 40.56 & 0.9749\\
    \cmidrule(lr){2-10}
    &\multirow{6}{*}{\begin{tabular}[c]{@{}l@{}}Sparse\end{tabular}}
    & \multirow{2}{*}{$\frac{1}{4}$}
    & Bilinear & 14.643 & 13.438 & 17.68 & 23.34 & 35.74 & 0.9579\\
    &&& Ours & 4.881 & \textbf{3.825} & \textbf{27.41} & 12.36 & 43.76 & 0.9878\\
    \cmidrule(lr){3-10}
    && \multirow{2}{*}{$\frac{1}{16}$}
    & Bilinear & 13.518 & 13.579 & 16.69 & 24.06 & 36.98 & 0.9690\\
    &&& Ours & \textbf{4.707} & 4.151 & 26.48 & 13.95 & 44.49 & 0.9897\\
    \cmidrule(lr){3-10}
    && \multirow{2}{*}{$\frac{1}{64}$}
    & Bilinear & 13.369 & 13.973 & 15.98 & 25.10 & 37.79 & 0.9732\\
    &&& Ours & 5.032 & 4.802 & 25.19 & 15.56 & \textbf{44.74} & \textbf{0.9902}\\
    \midrule
    
    \multirow{8}{*}{3.6}
    &\multirow{2}{*}{\begin{tabular}[c]{@{}l@{}}Conventional\end{tabular}}
    & \multirow{2}{*}{0}
    & - & 15.216 & 6.876 & 23.77 & 23.99 & 33.85 & 0.9233\\
    &&& Ours & 7.238 & \textbf{3.972} & \textbf{26.06} & \textbf{14.93} & 39.79 & 0.9685\\
    \cmidrule(lr){2-10}
    &\multirow{6}{*}{\begin{tabular}[c]{@{}l@{}}Sparse\end{tabular}}
    & \multirow{2}{*}{$\frac{1}{4}$}
    & Bilinear & 15.131 & 13.977 & 15.90 & 30.90 & 35.38 & 0.9493\\
    &&& Ours & 5.617 & 4.067 & 25.38 & 17.05 & 41.83 & 0.9759\\
    \cmidrule(lr){3-10}
    && \multirow{2}{*}{$\frac{1}{16}$}
    & Bilinear & 14.064 & 14.100 & 15.18 & 31.37 & 36.47 & 0.9592\\
    &&& Ours & \textbf{5.380} & 4.356 & 25.22 & 18.23 & 42.49 & 0.9789\\
    \cmidrule(lr){3-10}
    && \multirow{2}{*}{$\frac{1}{64}$}
    & Bilinear & 13.954 & 14.522 & 14.57 & 32.05 & 37.18 & 0.9630\\
    &&& Ours & 5.690 & 5.040 & 23.86 & 22.51 & \textbf{42.73} & \textbf{0.9798}\\
    \bottomrule
  \end{tabular}
  \label{tbl:comprehensive_eval}
  \vspace{-0.4cm}
\end{table*}

We use the PSNR and structural similarity index (SSIM) for the RGB images, root-mean-square error (RMSE) for the entire stokes vector $\mathbf{S}_{0,1,2}$ and polarization component, only $\mathbf{S}_{1,2}$ for polarization information, PSNR of the DoLP $\sqrt{s_1^2+s_2^2}/s_0$, and angular difference of the AoLP $\frac{1}{2}\tan^{-1}(s_2/s_1)$, $s$ represents the pixel value of $\mathbf{S}$.
We trained the model five times with different random seeds and used the average of the evaluation results.
The processing speed of our method was 0.05 seconds per $768 \times 576$ image.
Due to space limitations in the paper, a detailed evaluation has been provided in the supplementary material.

\vspace{-0.4cm}
\paragraph{Comparison with the basic methods (Sec.~\ref{sec:problem_formulation}):}
We first confirmed the effectiveness of our method (Eqn.~\ref{eqn:proposed}) by comparing it with  Eqn.~\ref{eqn:naive1} and Eqn.~\ref{eqn:naive2}.
Table~\ref{tbl:eval_naive} shows that for three different $r$ (1/4, 1/16, and 1/64), our method provides the most accurate results for polarization information.
Eqn.~\ref{eqn:naive2} has some good results, but the RMSE of $\mathbf{S}_{0,1,2}$ is extremely poor, indicating that $\mathbf{S}_0$ is not well estimated.

\vspace{-0.4cm}
\paragraph{Ablation study:}
Table~\ref{tbl:ablation} shows the ablation study confirming the validity of our network architecture described in Sec.~\ref{sec:network}.
Baseline is a method that compensates for four polarization intensity images.
The effectiveness of each measure—particularly, the polarization-constrained $\mathbf{S}_{1,2}$ compensation (SNA) and RGBRN—is verified.

\vspace{-0.4cm}
\paragraph{Comparison with depth completion networks:}
We also applied a general-purpose network (UNet~\cite{ronneberger2015u}, U2Net~\cite{qin2020u2}) and SOTA networks for depth completion and upsampling, GuideNet~\cite{tang2020learning}, NLSPN~\cite{park2020non}, and FDSR~\cite{he2021towards}, and compared their performance with our PCN in Sec.~\ref{sec:network}.
Each network was modified for polarization, and all conditions were identical except for PCN.
In other words, among the architectures proposed in Sec.~\ref{sec:network}, $\mathbf{S}_{1,2}$ compensation and RGBRN were adopted and the difference between our PCN and other networks was evaluated.
Table~\ref{tbl:network_comparison} shows the quantitative comparison results and Fig.~\ref{fig:result} shows the qualitative comparison of GuideNet.
Each result shows that our PCN can compensate for polarization information better than the other networks.

\vspace{-0.4cm}
\paragraph{Dataset (Sec.~\ref{sec:dataset}):}
We also evaluated the training performance of each of our datasets (real-world and synthetic), confirming the effectiveness of the large synthetic dataset, as shown in Tab. ~\ref{tbl:dataset_eval}.
The model can learn just as well from the synthetic dataset as the real-world dataset (with significant improvement in accuracy for $\mathbf{S}_{0,1,2}$) for the same number of training images. A higher number of training images can produce even higher quality results.
In addition, a mixture of real-world and synthetic datasets performs best because the domain gap is eliminated.

\vspace{-0.4cm}
\paragraph{Comprehensive noise-aware evaluation:}
To further investigate noise immunity, we reproduced situations of high noise factor $F_{\mathrm{n}}$ (high analog gain) from the noise model of the actual sensor and performed comprehensive evaluation of the performances of the conventional and sparse polarization sensors at various $r$, as shown in Tab.~\ref{tbl:comprehensive_eval}.
Observing the results of the evaluation, we see that our compensation method remains effective even in noisy situations ($F_{\mathrm{n}}=3.6$).
A visual representation of the evaluation results at $F_{\mathrm{n}}=0.72$ in Tab.~\ref{tbl:comprehensive_eval} is shown in Fig.~\ref{fig:trade-off}.
Compared to the conventional polarization sensor, our sparse polarization sensor obtains high-quality RGB images and polarization information, particularly for $r=1/4$ and $1/16$, with excellent balance between the quality of each output.
Furthermore, our compensation is effective in improving the quality of RGB images and polarization information for conventional polarization sensors.

\section{Discussions and conclusion}
We proposed a new sparse polarization sensor structure and network architecture that compensates for low-resolution polarization information to acquire high-quality RGB images and polarization information simultaneously.
The results of this study are based on simulations.
In the future, we intend to consider prototyping our spare polarization sensor to validate its practical application.
Based on this research, we believe that polarization pixels could be incorporated into many cameras in the future.

\vspace{-0.4cm}
\paragraph{Limitations:}
First, this study used a white color filter in the polarization pixel. Hence, we could not acquire wavelength-side polarization information.
Developing a sparse polarization sensor structure and compensation network architecture to acquire wavelength information is a topic for future research.
Second, cases where there is not a perfect correlation between the RGB image and polarization information, such as compensation in photoelastic measurements of transparent objects, are challenging to address with the current approach.
Finally, our network is not suitable for hardware implementation owing to the large number of parameters (184.8M).
We will work on making the network more compact in the future.

{\small
\bibliographystyle{ieee_fullname}
\bibliography{egbib}
}

\end{document}